\documentclass{article}

\usepackage{helvet}
\usepackage{PRIMEarxiv}

\usepackage[utf8]{inputenc}
\usepackage[T1]{fontenc}
\usepackage{hyperref}
\usepackage{url}
\usepackage{booktabs}
\usepackage{amsfonts}
\usepackage{amsmath}
\usepackage{amssymb}
\usepackage{nicefrac}
\usepackage{microtype}
\usepackage{lipsum}
\usepackage{fancyhdr}
\usepackage{graphicx}
\usepackage{float}
\usepackage{algorithm}
\usepackage{algpseudocode}
\usepackage{multirow}
\usepackage{subcaption}
\usepackage{mathtools}
\usepackage{cuted}
\usepackage{multicol}
\usepackage{caption}
\usepackage{natbib}
\usepackage{numprint}
\usepackage[table]{xcolor}
\usepackage{pifont}
\usepackage{comment}
\usepackage{makecell}
\usepackage{bbding}
\usepackage{enumitem}
\usepackage[symbol]{footmisc}

\DeclarePairedDelimiter{\nint}{\lfloor}{\rceil}

\DeclareMathOperator*{\argmin}{argmin}

\newcommand{\round}[1]{\ensuremath{\lfloor#1\rceil}}

\newcommand{\xmark}{\ding{55}}
\title{VLCQ: Post-Training Quantization for Deep Neural Networks
Using Variable Length Coding
\thanks{\textit{\underline{Citation}}:
\textbf{Abdel-Salam, R., Abdel-Gawad, A. H., and Wassal, A. G.}
VLCQ: Post-training quantization for deep neural networks using
variable length coding. Future Generation Computer Systems,
166, 107654, 2025.
DOI: \href{https://doi.org/10.1016/j.future.2024.107654}
{10.1016/j.future.2024.107654}.}
}

\author{
  Reem Abdel-Salam \\
  Computer Engineering Department \\
  Faculty of Engineering, Cairo University \\
  Giza, Egypt \\
  \texttt{reem.abdelsalam13@gmail.com}
  \\
  \And
  Ahmed H. Abdel-Gawad \\
  Computer Engineering Department \\
  Faculty of Engineering, Cairo University \\
  Giza, Egypt \\
  \texttt{ahamdy@eng.cu.edu.eg}
  \\
  \And
  Amr G. Wassal \\
  Computer Engineering Department \\
  Faculty of Engineering, Cairo University \\
  Giza, Egypt \\
  \texttt{wassal@eng.cu.edu.eg}
}

\begin{document}
\maketitle

\begin{abstract}
Quantization plays a crucial role in efficiently deploying deep learning models on resources constraint devices. 
Post-training quantization does not require either access to the original dataset or retraining the full model. Current methods that achieve high performance (near baseline results) require INT8 
fixed-point integers. However, to achieve high model compression by achieving lower bit-width, significant degradation to the performance becomes the challenge. In this paper, we propose VLCQ, which relaxes the constraint of fixed-point encoding which limits the quantization techniques from better quantizing the weights. Therefore, this work utilizes variable-length encoding which allows for exploring the whole space of quantization techniques. Thus, achieving much better results (close to or even better than the baseline results) while achieving lower bit-widths without the need to access any training data or to fine-tune the model. 
Extensive experiments were carried out on various deep-learning models for the image classification  and segmentation, and object detection tasks. When compared to state-of-the-art post-training quantization approaches, experimental results reveal that our suggested method offers improved performance with better model compression (lower bit-rate). For per-channel quantization, our method surpassed the FP32 accuracy and Piece-Wise Linear Quantization (PWLQ) method in most models while achieving up-to 6X model compression ratio compared to the FP32 and up-to 1.7X compared to PWLQ. If the model compression is the concern with little effect on performance, our method achieves up-to 12.25X compression ratio compared to FP32 within 4\% performance loss. For per-tensor, our method is competitive with Data-Free Quantization scheme (DFQ) in achieving the best performance. However, our method is more flexible in getting lower bit rates than DFQ across the different tasks and models.
\end{abstract}


\section{Introduction}\label{introduction}
Recent years have witnessed a rapid development of deep learning models in various tasks, such as  autonomous driving systems \cite{cui2021deep,lee2021deep,li2021deep}, object recognition \cite{qi2021review,bansal20212d,eitel2015multimodal}, object detection \cite{zhao2019object,wang2021salient,haciefendiouglu2021automatic} and text classification \cite{shaheen2020large,chia2019transformer,guo2020multi}. These models have achieved state-of-the-art results, at the cost of memory footprint, power consumption (computation complexity), and latency. These constraints limit their deployment on limited resource devices such as edge devices, embedded systems, mobile phones, and smartwatches.\\ In order to overcome this problem, several solutions have been proposed as searching and designing for small models\cite{wang2020apq,gong2019mixed}, compressing network by pruning redundant/unimportant parts in the network in order to accelerate inference \cite{wang2020apq,han2015deep} and network quantization which performs the computation in integer space \cite{lin2016fixed,wu2020integer}.

Most of the literature focuses on network quantization or pruning. These methods involve either training a model from scratch or fine-tuning a pre-trained full-precision model. Such methods are powerful as they improve network accuracy. One drawback of such methods is that they require access to a full training dataset, and many computation resources to perform backpropagation, which might not be feasible due to privacy issues related to the dataset, or resource constraints. Another drawback is that it is time-consuming, requiring a long time period for optimizing the model. A summary of the limitations of the literature
is shown in table \ref{comeptitors_downside}

\begin{table}[]
\caption{Drawbacks of the current state-of-the-art methods.}
\begin{tabular}{p{1.2cm}p{7.2cm}}
Methods                       & Limitations                                                                                                                                                                                                                                                                                                                                  \\ \hline
PWLQ       \cite{fang2020post}                   & \begin{tabular}[]{p{7.2cm}p{7.2cm}}Divides distribution into multiple regions, thus having higher bit-rate $b+1$, when having $2^b$ levels\\ (if b=8, the dense region will have 256 levels, while each of the sparse regions will have 127 levels which result in a total of 512 levels)\\ Uses a small dataset for activation calibration\end{tabular} \\ \hline
DFQ         \cite{nagel2019data}                   & \begin{tabular}[]{{p{7.2cm}p{7.2cm}}}Performance degrades beyond 6 bits\\ Achieves good results for per-tensor quantization only\\ Requires access to a small amount of data for high-bias absorption and bias correction steps\end{tabular}                                                                                                           \\ \hline
ACIQ     \cite{banner2019post}                      & \begin{tabular}[]{p{7.2cm}p{7.2cm}}Reports result for VGG16 and Resnet18 for Image classification task only.\\ Experiments were conducted for per-channel settings only\end{tabular}                                                                                                                                                                 \\ \hline
MultiPoint  \cite{liu2021post}                  & Approximate error by quantizing it and Adding it to original tensor                                                                                                                                                                                                                                                                          \\ \hline
\multicolumn{2}{p{8.2cm}}{AdaQuant \cite{hubara2020improving}, AdaRound  \cite{nagel2020up}, and BRECQ \cite{li2021brecq}: Need access to calibration dataset for fine-tuning}                                                                                                                                                                                                                                                             \\ \hline
\end{tabular}
\label{comeptitors_downside}
\end{table}

Therefore, it is important to quantize the models without access to the full original dataset. These types of methods are usually referred to as "Post-training quantization", which involve mapping weights and/or activation of the model from floating-point space (FP32) onto a regularly/irregularly spaced grid of fixed point in integer space (INT8), utilizing a scaling factor and an offset. This enables the use of quicker and more power-efficient integer operations in matrix multiplication and convolution computations, at the cost of reduced representational capability. Post-training quantization sometimes may require a small amount of the original dataset in order to fine-tune the model, usually for rounding problems, or scale selection. For a wide range of models, post-training quantization based on uniform INT8 quantization is sufficient to preserve near-original FP32 pre-trained model performance. However, when quantizing to lower integer space examples: INT7, INT6, INT5, and INT4, the performance degradation ranges
from very minor to catastrophic. This arises because the quantization error becomes significant enough to hurt the performance. The quantization error can aggravate the classification error due to ignoring layer dependencies, that quantizing a layer affects the next and the previous layers.
Most of the top performing post-training quantization schemes require access to a small amount of the dataset, to achieve high performance with low bit-rate. However, this might not be feasible in some cases where data is not available or sensitive as in the medical domain. On the other hand, post-training quantization without access to data suffers from performance degradation in low bit-width quantization, due to error accumulation from rounding schemes, fixed scale calculation, distribution shifts, etc. 
 \ref{comeptitors_downside} summarizes drawback of current work. In PWLQ the bit-rate is high thus having high performance, whilst having more information to store as which region, offsets, scales, and min-max. ACIQ is only tested on ResNet18 and VGG16 in a per-channel manner, leaving extra room for testing for other tasks and per-tensor schemes. In addition, it doesn't leverage from the underlying distribution. DFQ performs well in the per-tensor scheme however, performance degrades beyond 6 nit-rate quantization. Other methods assume access to calibration datasets.
 Hence, there is an urge for a quantization scheme, which should be able to quantize the model in per-tenor, and per-channel manner for different low bit-width while maintaining good performance on a variety of tasks and models. In addition to that to have a small memory footprint and take advantage of the underlying distribution. 
The aim of quantization is to decrease the model size and computation cost by performing operations in the integer space, thus decreasing the number of multiplication and add operations, while maintaining high performance. Hence, it could be deployed on mobile, edge devices which are battery, power, and memory constrained. In order to optimize the model furthermore, compression techniques could be applied on quantized weights to further decrease the model size and to allow for more compact representation. There are multiple ways to carry out compression fixed-length and variable-length coding. In a fixed-length coding, each value is represented as the same length, in terms of number of bits. In a variable-length coding (VLC), values may have different lengths. VLC is beneficial when the distribution of the possible values is non-uniform, thus representing the more common values with a lower bit rate, thus having an overall better compression, when compared with fixed-length code. For variable-length coding, there are different widely used algorithms such as: Huffman coding, Lempel–Ziv coding, arithmetic coding, and context-adaptive variable-length coding. Thus by leveraging VLC, deep learning models can be represented in a compact format, therefore reducing the storage required, without affecting accuracy or performance.


This paper focuses on post-training quantization for CNN models, without accessing the data or fine-tuning the models. In the absence of a dataset, the goal is to minimize quantization error, by taking advantage of the weights asymmetry, and scale selection to reduce rounding-off errors. We demonstrate significant improvements in quantization performance across a wide range of computer vision models that were previously thought to be difficult to quantify without fine-tuning. We utilize variable-length encoding as well as it helps in two aspects. First and foremost, it removes the fixed-length encoding constraint that limits all quantization techniques to be designed freely in order to reduce the quantization error. Without this constraint, our techniques in this paper can work in the general case where the distribution of the weights can be asymmetric and the number of bins can be as variable as needed to minimize the quantization error. These degrees of freedom allow better quantization of weights, thus achieving better performance. Second, in terms of bit-rate, it reduces the overall bit-rate since the  weights distribution is not uniform, but rather close to Gaussian or Laplace distributions.

VLCQ improves on existing work by having an assumption of having no access to the training datasets. Thus having much more generality. In addition, VLCQ takes advantage of the underlying asymmetry in distribution, thus dividing space into two regions only. This effectively decreases the number of information stored for the model and improves weights and activation quantization . In addition, conducting scale-selection or min-max search, is implicitly similar to the weight equalization step done in DFQ, which makes the model robust to tail-based distribution or outlier-based distribution. In addition it helps overcome rounding errors and improve performance for low-bit-rat quantization. This scheme is more general that makes the model have good performance in both per-tensor and per-channel quantization. In addition applying a mixed precision scheme on different granularities per-region, per-channel, and per-layer improves model performance while keeping the overall bit-rate at the desired level. Finally encoding model improves memory footprint compared to other approaches, and decreases bit-rate requirement of the model while having a large number of levels. Thus achieving good performance with low bit-rate. 

The main contributions of our work are as follows:
\begin{enumerate}
\item We propose post-training quantization with variable-length encoding (VLCQ): an efficient scheme to quantize weights and activation by taking advantage of weight asymmetry and variable encoding methods to further compress weights, thus achieving a low bit-rate, as shown in figure \ref{vlc_motive}.
\item We incorporate different scale selection methods and sensitivity metrics in order to overcome rounding-off errors due to quantization, without having access to any part of the dataset.
\item We apply mixed precision methods also to study their enhancements to VLCQ, which show significant improvements, and reduce quantization loss for low bit-rates.
\item We also experiment with non-uniform binning schemes to divide the space into different regions with different bit-widths.
\end{enumerate}
The rest of the paper is organized as follows: section \ref{sec:related_work} discusses recent methods, section \ref{sec:Methodology} discusses the proposed methods, section \ref{sec:Results} shows experimental results, and section \ref{sec:Conclusion} concludes the paper.

\section{Related Work}
\label{sec:related_work}
There has been a great effort in accelerating efficient deployment (in terms of computational power, memory, or energy consumption) of deep learning models on low computational devices. This attracted research into the area of model compression. There are a variety of approaches to achieve that.
The first group relies on designing networks with efficient layers. Examples include depth/point-wise layers \cite{hua2018pointwise,he2019depth}. The second group relies on optimizing the network architecture either through network pruning or activation compression, knowledge distillation, or quantization.
{The work in \cite{yang2024streamlined} introduces the EL-YOLO algorithm, an optimized version of YOLOv8, designed for efficient maritime object detection using RGB cameras. It features novel enhancements such as AWIoU, SMFN, and GDFP, which improve convergence speed, detection accuracy, and network efficiency, respectively. AWIoU improves object localization by assigning different weights to training examples based on quality, reducing the impact of low-quality samples. SMFN enhances the neck component by adding scales and shortcut connections to better analyze features, especially for small objects in maritime settings. GDFP employs greedy filter pruning to reduce the network size by removing less important filters while maintaining accuracy. Together, these improvements significantly boost EL-YOLO’s performance in challenging maritime object detection scenarios.}
\\

Model quantization refers to mapping model weights and/or activation from float space to integer space. Model parameters can be stored in fewer bits, and computations can be carried out on integer units rather than power-hungry floating-point ones. For model quantization, the literature can also be divided into two categories based on the expense and complexity of the quantization procedures: post-training quantization, and quantization-aware training. \\

Quantization-aware training: This method works on re-training the model from scratch by simulating low-precision behavior (integer space) in the forward pass, while the backward pass remains the same. This makes model parameters more robust to quantization making the whole process almost lossless. The authors in \cite{tailor2020degree} proposed a quantization scheme  DEGREE-QUANT (DQ) to overcome quantization degradation in graph neural networks. The authors identified the main cause of errors which is rounding error, especially at nodes with high in-degree. In order to overcome this, DQ works by protecting high in-degree nodes from quantization thus encouraging more accurate gradients to flow back to the weights.
\\
Post-training quantization (PTQ): These methods work on quantizing a pre-trained model without the need to re-train it. PTQ can be further divided into subcategories: data-free quantization, mixed-precision quantization, or quantization with a small calibration dataset.
\\

Data-free quantization: refers to the quantization of a network without any knowledge of the training dataset used for training the model.
The work in \cite{banner2019post} proposed the first practical 4-bit post-training quantization approach. The proposed methodology relies on three steps: Analytical Clipping for Integer Quantization (ACIQ), per-channel bit allocation, and bias correction. In Analytical Clipping, the best minimum and maximum are chosen in order to clip activation values to mitigate errors. In per-bit allocation, each channel in the convolution layer has a different b-bit-width allocated, but the average bits are equal to require bit-width d. As a result of quantization, the new distribution of weights is much different than the original one, to overcome this bias correction is used which shifts the mean of the quantized weights near the original full precision weights and rescales the quantized distribution in order to have a variance similar to the original one. 
\\
Piece-Wise Linear Quantization (PWLQ) \cite{fang2020post} is a scheme to quantize networks that have a significant model degradation with low bit-widths. PWLQ works on dividing model weights and activation into three non-overlapping regions: a dense region (center around zero) and two sparse regions. The dense region is quantized with b-bits, while the two sparse regions are quantized with b-1 bits. Therefore the total number of levels is doubled (if b=8, the dense region will have 256 levels, while each of the sparse regions will have 127 levels which results in a total of 512 levels).
 In addition to that, they need an extra bit for storing the region in order to determine which region the weight value belongs to, so that re-quantization (dequantization) could work.
The research work in \cite{nagel2019data} proposed a Data-Free Quantization scheme (DFQ) for per-tensor weight/activation quantization. DFQ relies on cross-layer equalization, which re-scales each channel in the convolution layer weights/activations so that the overall minimum and maximum in each channel are similar, hence when choosing a scale, all the values would have a corresponding value in the integer space other than zero. In addition to cross-layer equalization, DFQ relies on absorbing bias introduced by the quantization procedure. However, such a scheme shows its effectiveness only in per-tensor quantization only (In per-tensor quantization is applied to the whole tensor irrespective of its shape.). However, DFQ does not work well in per-channel (In per-channel quantization the weights of a convolution layer tensor are quantized for each output channel independently.) quantization as per-tensor quantization. Furthermore, it cannot be integrated with other methods such as PWLQ, or others.

 Generative low-bit-width Data Free Quantization \cite{xu2020generative} is a data-free quantization scheme, which relies on similar concepts as \cite{choi2020data}, where three models are used: full precision model, generator model to generate synthetic data, and the quantized model. GDFQ relies on generating synthetic data from the generator model by
learning knowledge from the pre-trained full-precision about classification boundary knowledge and data distribution. Then this data is used to fine-tune the weights of the quantized model. This scheme shows its effectiveness for low bit-width for weights and activation. Post-training quantization with a small calibration dataset refers to those methods which have access to a small subset of the training data. These methods use data to either fine-tune the quantized model or search for a better scale.  

Quantization with a small calibration data set: refers to the quantization of a network while having access to a small subset of the dataset. This could help in fine-tuning the model, or adjust some hyperparameters as minimum and maximum when doing quantization for activation.
The authors in \cite{nagel2020up} proposed Adaptive Rounding technique for post-training quantization (AdaRound). Their motivation is based on the fact that rounding to the nearest integer is not always optimal. It may be optimal from the perspective of Mean Squared Error (MSE), but generally, it might not be good from the point of view of activation error. AdaRound performs per-layer fine-tuning with a small calibration dataset, in order to find the best rounding option for each weight value, by minimizing activation error. Similar to AdaRound, the authors in \cite{hubara2020improving} proposed AdaQuant. AdaQuant is quite similar to AdaRound, but the authors introduced three other modifications: 1- per-layer bit-allocation using integer programming, 2- batch-norm tuning after quantization, and 3- bias tuning. 
The research work in \cite{li2021brecq} proposed a PTQ scheme using block reconstruction (BRECQ). BRECQ is quite similar to AdaRound, but the authors introduced two modifications: 1- initial scale selection by adjusting the minimum and maximum of distribution and choosing the one with the lowest Lp-norm loss. 2- Instead of calculating activation error per-layer and adjusting rounding based on it, which neglects interaction between layers, BRECQ divides the model into blocks, and minimizes activation error per-block, thus taking into account layer interactions in the same block.

Mixed precision quantization: refers to those methods that allocate different numbers of bits for each layer or each channel. However, a major problem is how to determine the sensitivity and importance of each layer to allocate the corresponding bit-width for quantization. The research work in \cite{chen2021towards} formulated the problem of allocating/assigning bit-width for each layer as a discrete constrained optimization problem. However, using the original training objective (loss function between predicted using the quantized model and original labels) makes the problem intractable. In order to solve it, the authors used  Taylor expansion to approximate the objective and provided an efficient approach to calculate the Hessian matrix. Finally, they showed that the problem could be reformulated as a Knapsack problem called Multiple Choice Knapsack Problem (MCKP) and proposed a greedy search algorithm to solve it efficiently. The authors in \cite{chikin2022data}  introduced a new network compression method called PNMQ. It does not rely on expensive calculations like gradient descent training or clustering. The method involves using a parametric family of non-uniform quantization grids that can be easily optimized. These grids improve the quality of quantization approximation compared to uniform quantization. The paper also presents an optimization pipeline for tuning the non-uniform grids for each model layer, with or without using data. Additionally, the authors demonstrate that different layers of models require different bit-widths for quantization approximation and propose a data-free algorithm for selecting the appropriate bit-widths based on comparing quantization errors. This algorithm automatically selects optimal bit-widths to achieve a user-specified compression ratio. PNMQ can be applied to any model without changing the network structure and is compatible with existing weight compression techniques. 

\section{Motivation} 

The goal of PTQ is to successfully quantize a deep learning model, with small amounts of performance degradation, by transferring floating point values to integer values and carrying out operations in integer space. The quantization process could be carried out while accessing a small amount of dataset, or not accessing any amount of it. Many trained FP32 deep learning models could be quantized to INT8 without much loss in performance using per-channel or per-tenor quantization. Some models as MobileNet family incur loss in performance varies up to 5\% performance degradation  due to the nature of its layer decomposition. MobileNet family are mainly decomposed of a variant from convolution layers which are pointwise and depthwise separable layers. The performance keeps degrading for most of the CNN families such as ResNet, MobileNet, Inception, etc when we quantize using lower bit-rate, i.e, apply INT5, INT4, INT3, and INT2 quantization to these models.  The degradation arises due to multiple factors. First, there is a distribution shift between the quantized model and the original FP32 model. Second, the nature of the rounding operation in the quantization operation induces noise. Third, neglecting the per-layer or per-block interaction while quantization.
A typical CNN structure (adapted from \cite{radosavovic2020designing}) consists of a stem layer (initial convolution on input images), a body layer, and a head layer (average pooling with a fully linked layer). A body is made up of numerous stages, and each stage is made up of several blocks, each block is composed of basic layers as convolution, maxpool, etc. Error in one layer mitigates and affects subsequent layers. 
Finally, the error metric to measure noise is based on weights not original task loss or activation. To overcome such degradation and problems, per-channel quantization with fine-tuning is usually applied. However, this might not be practical in some cases where a small portion of the dataset might not be accessible due to privacy concerns or license issues. Therefore we cannot solely rely on those methods.

\subsection{Weight ranges and distribution} 
Most of the literature focuses on per-channel quantization for one main reason. It yields higher performance for quantized models when compared to per-tensor quantization at the expense of more storage will be needed to store the extra scale for each channel. The reason for better performance lies in the nature of the weight distribution for each channel in the model. According to the work in \cite{blundell2015weight}, all deep learning model weights and activation distribution follow a bell-shaped distribution similar to the Gaussian or Laplacian distribution. However, distribution shape varies according to model architecture or task. Some models have exact Gaussian or Laplacian distributions. Others might have a long-tail Gaussian distribution or have outliers.

\begin{figure*}[t]
\begin{subfigure}[b]{0.3\textwidth}
    \includegraphics[width=\textwidth]{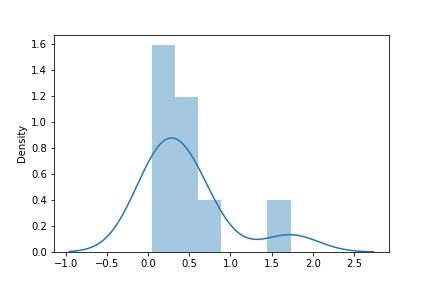}
     \caption{The  weights distribution lies entirely in the positive space having values ranging from [0,15] }
    
     \end{subfigure}
    \begin{subfigure}[b]{0.3\textwidth}
    \includegraphics[width=\textwidth]{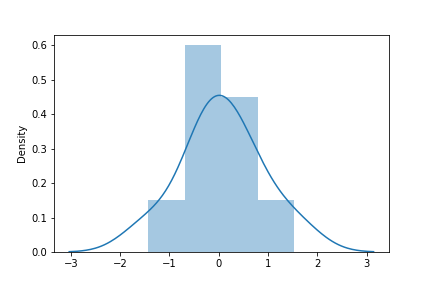}
    \caption{The  weights distribution lies in both positive and negative space ranging from [-1.5,1.5] with zero mean}
     \end{subfigure}
    \begin{subfigure}[b]{0.3\textwidth}
    \includegraphics[width=\textwidth]{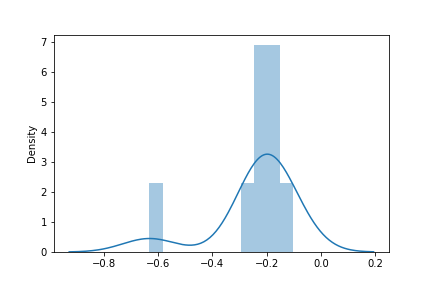}
     \caption{The  weights distribution lies entirely in the negative space having values ranging from [-0.1,-0.6] }
     \end{subfigure}
\caption{The weights distribution of selected channels in the first convolution layer in the 10th Inverted Residual} block, in MobileNetV2 model.  
\label{per-output--channel-weight-dirstibution-mobilenet-v2}
\end{figure*}

Some models have high variations in the ranges of weights for each channel in a given layer. As shown in figure \ref{per-output--channel-weight-dirstibution-mobilenet-v2}, weight ranges vary drastically from channel to another even in the same layer as in MobileNetV2. Some channels have a symmetric range of [-1.5, 1.5], while others have asymmetric and skewed range [-0.1, 0.6]. If per-tensor quantization is applied on the weights, the weights which have small ranges will be clustered in very few bins and many of them might get mapped to 0, thus degrading the performance. However, this gain in using per-channel quantization is not applicable in some models as DeepLab-v3+
with a backbone of MobileNetV2, where several layers of weight ranges become very large (e.g., [-54.4, 64.4]). Thus while applying per-channel quantization, the model still suffers from severe performance issues. In addition to per-channel quantization, uniform quantization is applied, which assumes symmetry and uniformity of levels in the underlying distribution.  In practice, the weight and activation distributions of pre-trained DNNs are similar to bell-shaped Gaussian or Laplacian distributions. As a result, uniform quantization may not always produce a small enough approximation error to preserve model accuracy, particularly in low-bit cases. 

\subsection{Quantization error} 
Uniform symmetric quantization converts floating-point numbers to a set of fixed-point values. These points (or grids) are symmetrically distributed and have the same interval. For an unsigned integers, the symmetric quantizer maps the values to be in the range of $ \{0,...., 2 ^{b}-1\}$ for a specific  bit-rate b, while in the range $ \{-2 ^{b-1},...., 2 ^{b-1} -1 \}$ for signed integers. 
The uniform quantization scheme at b-bits can be defined as:
\begin{equation}
\label{eq1}
\begin{aligned}
& \text{Uniform\_Symmteric\_Quantization}(b, w):  \\
& \quad   \beta=\max(\text{abs}(w_{min}), w_{max}) \\
& \quad   \delta= 2*\beta/(2^b - 1) \\
& \quad   w_q= \nint{w/\delta}  \\
\end{aligned}
\end{equation}
where $b$ is the desired bit-rate, $w$ is the weights or activations of the model to be quantized, $\delta$ is the scaling value or the step size, and $w_q$ is the new values quantized.  The function converts a real number to an integer value using a rounding procedure (e.g., round to nearest or truncation). This function is essentially a mapping from real numbers to some integer value. It is  possible to recover the values back in the real space (floating point space) through an operation that is usually called dequantization:
\begin{equation}
\label{eq4}
\begin{aligned}
Deq(w_q,\delta): w_{deq} = w_q* \delta
\end{aligned}
\end{equation}
where $Deq$ is the dequantization function $w_{deq}$ is the dequantized values (representation of the values in the real space), $w_q$ is the values represented in integer space, $\delta$ is the scale. \\
In quantization, the goal is to reduce the precision of the parameters ($W$) as well as the output activation maps of each intermediate layer with minimal degradation in the performance of the model (in terms of accuracy, speed, and memory).
The goal can be formulated as either one of the following two objectives, or a weighted sum of both:
\begin{equation} \label{eq2}
\begin{split}
 \min \quad & || W - W_{deq}||^2, \quad \text{or} \\
 \min \quad & || W * X - W_{deq}*X_{deq}|| \\
 \text{s.t.} \quad &  W_{deq},\;X_{deq} \in \text{dequantized space}
\end{split}
\end{equation}
where $W$ is the original weight in floating point space, $X$  denotes output activation. $W_{deq},X_{deq} $ are dequantized weights and activation. The first objective is minimization problem, and could be achieved by using a rounding operator to weights. However, new empirical or theoretical evidence \cite{nagel2020up,li2021brecq,hubara2020improving,nahshan2021loss} suggests that just reducing quantization error in weight's space does not result in optimal task performance.  On the other hand, minimizing the latter objective requires access to a small amount of the dataset, to carry out fine-tuning. The fine-tuning is carried out to either optimize quantized weights values with respect to original task loss, or activation error. New methods  \cite{nagel2020up,li2021brecq} suggest that fine-tuning could be carried out to decrease quantization error due to non-optimality of the rounding operator. The fine-tuning is carried out to choose the best rounding strategy with respect to task loss. However, these methods require access to a small amount of dataset around 1024 images, which in some problems might not be feasible to have.  

\begin{figure*}[h!]
\centering
\includegraphics[width=0.7\textwidth]{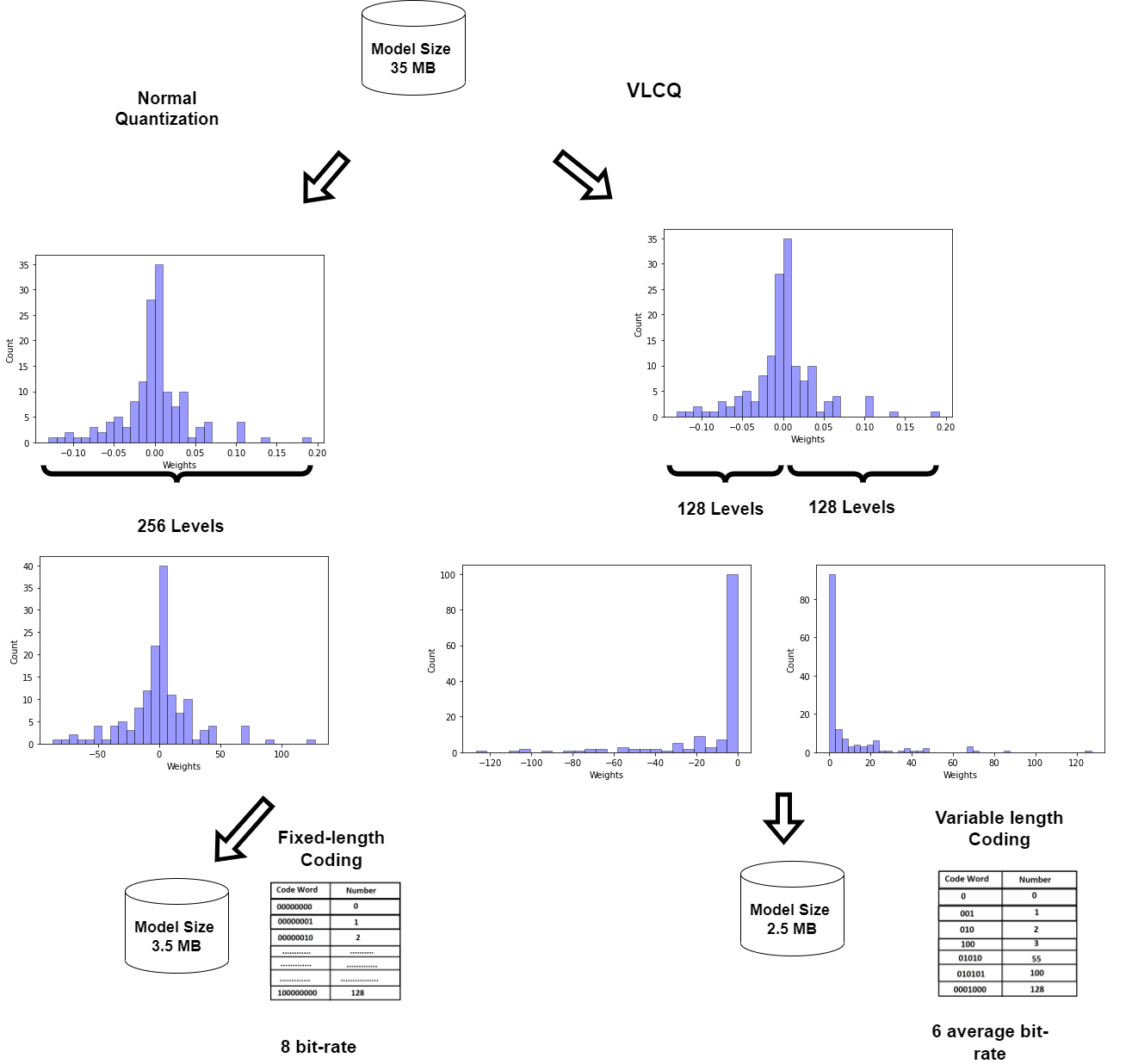}
\caption{VLCQ motivation compared to uniform quantization. }
\label{vlc_motive}
\end{figure*}

Another problem is in the quantization operator itself. Most of the parameters used are fixed such as scale, minimum and maximum values. However, these parameters could be learned or explored further to study their effect in quantization and how they affect performance whether positively or negatively. \\

Therefore, there is a need to have a framework which could quantize a model to perform well in per-tensor or per-channel modes for high and low bit quantization while taking advantage of the weights distribution. In addition, a work-around for rounding schemes in absence of dataset. Finally, study the effect of different hyper-parameters on the quantization operator and their correspondence.

Figure \ref{vlc_motive} shows the whole process of contrasting our base method, VLCQ, to the normal uniform quantization scheme.

\section{Methodology}

In this section, we start by discussing the proposed quantization method which relies on utilizing 2-scales quantization \ref{2scales}. This method leverages from the asymmetry of the weights distribution by dividing the distribution weights into two regions and applying a quantization operator in each region. This is the core method that we expand to have more variant methods. One variant relies on utilizing Mixed-Precision quantization \ref{MP} with the 2-scales core method by applying it at multiple levels such as per-channel or per-region. The motivation behind it is that not all channels have the same importance, thus by allowing some channels to have more bits/levels, the quantization performance might improve. In addition, some channels might have large values of positive numbers while small values of negative values or vice versa. Thus allocating a fixed number of bits (8-bits, 4 bits..etc) for the channel might not be a good choice. Therefore we allow more bits/levels to represent a certain region than the other.  A second variant is through studying the impact of non-uniform quantization by allowing multiple step sizes. In a third variant, we move forward toward studying the impact of scale choice in the quantizer \ref{ss-min-max} operator by allowing exploration of different scales and how it affects activation  and weight error. In the last variant, we explore further impact of minimum and maximum distribution in quantizer operator and how they affect quantization noise. Finally, variable length Golomb code \ref{vlc}
 is discussed to allow compact representation for the quantized model across all variants we propose.

\label{sec:Methodology}
\subsection{Post-training quantization using variable length coding} 
\label{2scales}

In this section, We discuss our proposed methods of post-training quantization with variable-length encoding (VLCQ) and demonstrate that it has a greater representational capability (lower quantization error in low-bit widths) compared to other schemes.
In order to improve model accuracy due to quantization, one needs to account for an accurate approximation of the original model's  weights distribution. We follow this criterion to investigate different quantization schemes, even though the relationship between quantization error and model accuracy is not exploited yet, without knowledge of the original dataset. \\

To handle the general case when the distribution is asymmetric, our approach is based on dividing the distribution into two regions, if applicable. The first region is the space of negative values and the second one is the space of positive values. For each region, the number of levels assigned for quantization can be both equal to $N=2^{b-1}$ bins or both are different based on the methods we describe later.

\begin{figure}[t]
\centering
    \begin{subfigure}[b]{0.28\textwidth}
    \includegraphics[width=1.15\textwidth]{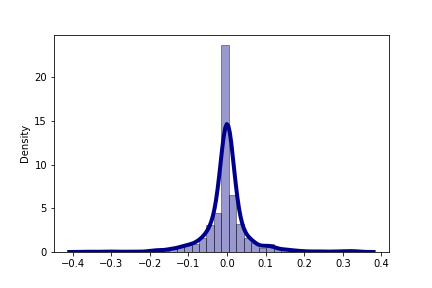}
    \caption{The weights distribution for the first convolution layer. The weights are centered around zero, while the variance ranges from [-0.4,0.4] }
     \end{subfigure}
    \begin{subfigure}[b]{0.3\textwidth}
    \includegraphics[width=1.04\textwidth]{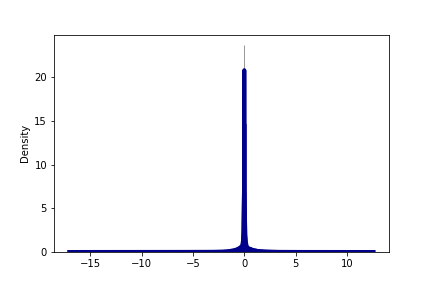}
     \caption{The weights distribution for the second convolution layer, in the 10th block. The weights are centered around zero, while the variance ranges from [-15,11] }
     \end{subfigure}
\caption{The weights distribution in a different layer in MobileNetV2 model. The mean of the distribution is almost centered around zero while the minimum and maximum of the distribution range for different convolution layers. }
\label{per-tensor-weight-dirstibution}
\end{figure}

For break-point estimation, we either use zero or the median of the distribution as our break-point to divide our space into one or two  regions according to the distribution underlying. In the case of per-tensor quantization, as shown in figure \ref{per-tensor-weight-dirstibution}, most of the distributions are symmetric around zero, hence we divide our space into 2  non-overlapping regions, the positive region $R_+ = [0, W_{max}]$, where $W_{max}$ is the maximum value in the vector of weights $W$ to be quantized, per-channel or per-tensor, and the negative region $R_- =[W_{min}, 0)$, where $W_{min}$ is the minimum value in the weights $W$.

In each region, uniform symmetric quantization is applied in such a way that every value in the quantization range, including the sign, is represented in the desired bit-width b. We define the
desired bit-width b quantization scheme as in Equation \ref{eq2}.
\begin{equation}
\label{eq2}
\begin{split}
Q(W^{(c)},b,S) &= \begin{cases}
uq(W_i^{(c)},b-1,R_-,S) & \text{if} \; W_i^{(c)} \in R_- \\
uq(W_i^{(c)},b-1,R_+,S) & \text{if} \; W_i^{(c)} \in R_+
\end{cases} \\
S &= \frac{(W_{max}-W_{min})}{(2^b -1)}
\end{split}
\end{equation}

where $uq$ is the uniform quantization function, that maps floating point values into lower integer space representation, $b$ is the desired bit-width and $W^{(c)}$ is the vector of weights for channel $c$, in case of per-channel quantization. The same equation holds for the outputs of the activation function. The uniform quantization function for symmetric quantization and  desired bit-width $b$ can be defined as

\begin{equation}
\label{eq4}
uq(x_i,b,R,S)= \lfloor{\frac{x_i}{S}} \rfloor 
\end{equation}

where $x_i$ is the value to be quantized, and $R$ is the given range to quantize for.

In the case of per-output channel quantization, as shown in figure \ref{per-output--channel-weight-dirstibution-mobilenet-v2}, not all weights distribution can be divided into positive or negative regions. There are cases of distributions that either lie entirely in the positive space or in the negative space. In such cases, it has been shown in the experiments that dividing them will not affect the model performance. Hence for such cases, one scale symmetric quantization is applied, which uses the whole bit-width $b$ (i.e all levels in the desired bit-width are allocated for that region). Equation \ref{eq3} extends Equation \ref{eq2} to add this case of the whole range of weights $W^{(c)}$ contained in either $R_-$ or $R_+$,
\begin{equation}
\label{eq3}
Q(W^{(c)},b,S)=\begin{cases}
$uq$(W_i^{(c)},b-1,R_-,S) & \text{if} \; W_i^{(c)}\in R_-\\
$uq$(W_i^{(c)},b-1,R_+,S) & \text{if} \; W_i^{(c)}\in R_+\\
$uq$(W_i^{(c)},b,[W_{min},W_{max}],S) & \text{if} \; W^{(c)} \subseteq R_- \;\text{or} \; W^{(c)} \subseteq R_+.
\end{cases}
\end{equation}
Therefore, we propose 2 scales quantization schemes that leverage from the asymmetry of the underlying distribution. In order to quantize weights, the distribution is divided into two halves around zero. Then each half is quantized using the same number of levels $2^{b-1}$. However, this is not optimal in terms of approximation, due to the asymmetry. Therefore it is better to relieve this condition by allowing different halves (regions) to have different numbers of levels accordingly. This leads to different decision criteria for level number selection, which is discussed in the next subsections. 

\subsection{ Mixed precision Quantization}
\label{MP}
In mixed-precision quantization, each layer or channel in the model is assigned a different number of bins, thus utilizing the asymmetry of the distribution of the weights to redistribute the bins across.
However, mixed-precision quantization is difficult for many reasons. First, it's difficult to estimate the sensitivity of each layer or channel without accessing the full dataset. Therefore choosing random bit-width $b$ for each channel or layer might result in performance degradation. Second, the search space for choosing the number of bins for each channel or layer is large, and an exhaustive search within each layer as well as across all layers is infeasible. \\
In our approach, we focus on per-channel mixed-precision quantization. Instead of limiting all channels to have the same number of bins/bits, some channels are allowed to have higher or lower numbers.  The only requirement we have is that the total bit-rate remains the same or less than the required. We further use mixed-precision per region. For a specific channel, symmetry does not often exist (if we divide the distribution into two regions around zero positive space and negative space). Therefore some regions may require more levels than other regions. For example, we could find a channel where the total unique negative values are 16, and the total unique positive values are 500. Hence, to better approximate the distribution, the number of levels allocated in positive space should be larger compared to negative space. 

The techniques for mixed precision quantization that are applied:
\begin{itemize}
  \item  Non-uniform quantization through multiple break-points. Further breaking each region into two parts dense part and sparse part using a breakpoint (the point at which divide sparse/ tail of the distribution and dense part ) and applying non-uniform quantization at each subpart.
  \item  Per-region bit-allocation.
  \item  Per channel per-region bit-allocation.
 \end{itemize}
\paragraph{Non-uniform quantization through multiple break-points} \hspace{0pt} \\

\begin{figure}[h]
\centering
\includegraphics[width=0.9\linewidth]{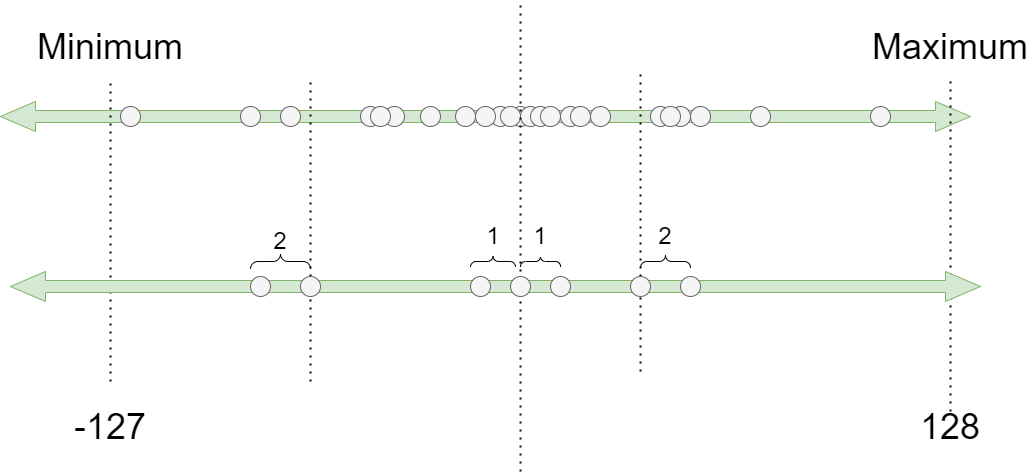}

\caption{Non-uniform quantization through breakpoints and variable step size.}
\label{NOn-uniform-quant}
\end{figure}

In this approach, we adopt the idea of multiple break-points in each region similar to \cite{choi2020data}, but instead of having the same number of bits or levels or bins in each sub-region, the number of bins varies by a factor of 2, as shown in figure \ref{NOn-uniform-quant}. The first arrow represents the distribution in the FP32 space, the green dots represent the weights as indicated, the region around zero is dense, while the tail of distribution contains less values. The second arrow indicates quantized weights in INT8 space; the region around zero has a step size of 1, while sparse regions starting from dashed vertical line have a step size of 2, since not all levels are needed due to the sparsity of distribution. At first, the distribution is divided into two regions (if available): a negative region and a positive region. Then each region is subdivided into 2 more regions where the tail has double the bin width. This is because more accuracy is needed near the center in contrast to the tails, thus the bin width is small close to the center, and becomes wider at the tail. If we consider that $n_1$ and $n_2$ represent the number of bins for the first and second regions respectively, where the first region represents the region from 0 to the break-point $p$ and the second region from $p$ to the end of the range. If the number
of bins is $B$ which includes the zero bin and two equal numbers of bins for the positive and negative ranges, then $n_1 + n_2 = (B-1) / 2$. We ensure that the width of the bins is $x$ and $2x$ in regions 1  and 2 respectively. Starting with the set of Equations in \ref{eq1}:

\begin{equation} \label{eq1}
\begin{split}
n_1 + n_2 = (B-1)/2 \\
n_1 x + n_2 (2 x) + \frac{x}{2} = R\\ 
n_1 x + \frac{x}{2} = p, \\
\end{split}
\end{equation}

Therefore, $n_1$ can be estimated as in Equation \ref{eq1_f},

\begin{equation} \label{eq1_f}
n_1 = \frac{p(B-1) - (R-p)/2}{R+p}.
\end{equation}
$R$ can be computed in one of three ways: $(|W_{min}|+W_{max}) / 2$, $\min(|W_{min}|, W_{max})$, or $\max(|W_{min}|, W_{max})$. In the case of average or min, we need to perform clipping.

\paragraph{Per-region bit-allocation and Per-channel per-region bit-allocation} \hspace{1pt} \\
In this approach, we adopt the same method proposed by \cite{banner2019post} to calculate sensitivity with some modifications.

In order to determine the assigned number of bits for each channel or region the following Equation is used: 
\begin{equation}
M_i= \round {\log_2{(\gamma^{2/3}/ (\sum_{n=1}^{N}\gamma^{2/3} ))}.B}
\end{equation}
where $M_i$ is the number of bits assigned for a specific channel or region, $N$ is the number of channels for per-channel bit-allocation or number of regions which is two in the case of per-region bit-allocation,$B$ is the average desired bit-width. 
We propose several ways to calculate $\gamma$: 
\begin{itemize}
  \item The standard deviation of channel or region.
  \item  The number of unique values for a given channel or region.
  \item The standard deviation of a channel or region weighted by the number of unique values for a given channel or region.
\end{itemize}
For per-region bit allocation $\gamma$ is the standard deviation of the region. Per-channel per-region bit-allocation, the first time for calculating the number of bits for the channel $\gamma$ is the standard deviation of the channel, while when calculating per-region value $\gamma$ is the number of unique values in the region. In both cases, we limit the space of the allowed bit-width to be between $[3-8]$.

\subsection{Scale selection}
\label{ss-min-max}

In PTQ, usually weights are quantized by mapping FP32 values to the nearest integer values, $W_q= round(w/s)*s $, which can be formulated as $W_q=dequant(quant(W))= W+n_w$, where $n_w$ is the noise introduced and can be modeled as a shift in weights.
Often, this is known as rounding to the nearest integer strategy. However, as demonstrated by \cite{nagel2020up}, it's not always optimal to round to the nearest. Rounding to the nearest may produce a small mean square error (MSE) with respect to original weights, However, it does not mean that it will produce a small error with respect to output activation or task loss. Table \ref{table:rounding_techniques}, shows the performance of the model when the whole network is quantized using different rounding techniques. It can be concluded that the best scheme was flooring. 
Rounding to the nearest introduces bias and variance shifts in the quantized weights. In order to overcome this issue, bias and variance correction can be used as described by \cite{nagel2019data,banner2018aciq} to decrease the rounding error effect. This problem is partially solved for high bit-width as 8,7,6 bit quantization, however, for low bit-width quantization as in 5,4,3 bits quantization, this problem still exists.
In order to mitigate this problem, we tackle this problem from another perspective. Instead of finding an optimal rounding strategy as described by \cite{nagel2020up,li2021brecq}, We employ three techniques:
\begin{itemize}
  \item  Exploring search space of possible scales (Scale search / Fraction bit-width).
  \item  Finding the best scale by changing the minimum and maximum of the weights (Min-Max adjustment).
  
  \item  Exploring best minimum and maximum space with scale search for each combination.
  
  \end{itemize}

\begin{table}[h]
\caption{Comparison of ImageNet validation accuracy, MSE of weights and activation among different rounding schemes, for INT8 uniform quantization using MobileNetV2 model}
\centering

\npdecimalsign{.}
\nprounddigits{3}
\footnotesize
\begin{tabular}{@{\extracolsep{\fill}}>{\centering}m{50pt}>{\centering}m{50pt}>{\centering}m{50pt}>{\centering}m{40pt}}
\toprule[1.5pt]
Rounding Technique  & MSE of Weights      & MSE of Activation     & Accuracy \tabularnewline \toprule[1.5pt]
Nearest & \numprint{0.16068011}  & \numprint{0.0074790} & \numprint{71.752}\%   \tabularnewline \hline
Floor               & \numprint{0.1150858} & \numprint{0.0058277}  & \numprint{71.814}\%    \tabularnewline \hline
Ceil                & \numprint{0.2391723}   & \numprint{0.0092739}  & \numprint{71.752}\%   \tabularnewline
\bottomrule[1.5pt]
\end{tabular}
\npnoround
\label{table:rounding_techniques}
\end{table}

In the first approach, we try to find the scale that minimizes the Lp norm with respect to weights. Scale, in essence, plays a crucial part in quantization as it determines the new centers,
in which the values need to be mapped. When the weights are divided by a scale to obtain integer values, the results are not purely integer values, but rather still in the floating space. Thus, these values need to be mapped to an integer level whether up, or down if not exactly an integer value. Each decision shifts the distribution of weights thus model performance changes. Therefore choosing appropriate scales helps in reducing rounding errors. In order to do so, different ranges of scales need to be investigated for a specific bit-rate.
Scale search strategy partially solves the problem without knowledge of the data. Finding the best scale that minimizes the Lp norm is similar to the rounding solution. The space of possible scales to achieve a certain bit-width is infinite. If we consider what each scale does, it remaps weights to new centers (re-bin weights to specific centers based on a given scale). Therefore each scale in the infinite space differs in the new centers of the weights. In other words, each scale performs different rounding on the weight, therefore our goal to partially solve rounding errors can be defined as finding the best scale that minimizes rounding errors. However, it's not feasible to search this infinite space. To overcome this problem, The search space is restricted to be the interval of possible scales that lies between the scale of the first higher bit-width and the scale of the lower bit-width. This interval is divided equally based on specified steps, chosen according to the type of quantization and bit-widths. This criterion holds as we decrease bit-width, the search space is double the size of the previous one.
For a fixed minimum and maximum we compute different scales by adjusting a number of levels (having fraction levels example: 128.5 levels or 129 levels or 127.5 levels) or by directly exploring the space between the first higher bit-width and the scale of the lower bit-width.
The pseudo-code of the whole algorithm is presented in Algorithm \ref{alg:SearchSpaceScale}.
\begin{algorithm}[h]
\caption{Semi optimal Scale selection}
\label{alg:SearchSpaceScale}
\begin{algorithmic}

\State{\textbf{Input:} Weights of tensor (W), bit-width (b)}
\State{\textbf{Output:} Scale (s), quantized weight tensors (Wq)}
\State{ \underline {\textbf{Choose optimal P values for Lp norm }}}

\State \texttt{$S_0= (W_{max}-W_{min})/ (2^b-1)$}
\For{\texttt{p in [p1.......pk]}} 
        \State \texttt{$P_{optimal}= \argmin_{p}((||Q(W,b,S_0)-W||^p))^{1/p}$} 

\EndFor
\State{\underline{\textbf{Scale Selection}}}
\State \texttt{$S_{large}= (W_{max}-W_{min})/ ((2^b+1)-1)$}
\State \texttt{$S_{small}= (W_{max}-W_{min})/ ((2^b-1)-1)$}

\For{\texttt{$S_{candidate} \in [S_{large}.......S_{small}]$}} {
 \State \texttt{$Wq= Q(W,b,S_{candidate})$} 
 \State \texttt{$loss=lp\_norm(P_{optimal},Wq,W)$} \State \textbf{\If{$current\_loss > loss$}
 \State \texttt{$S_{optimal} = S_{candidate}$}
 \EndIf
  }
}\EndFor 
\end{algorithmic}
\end{algorithm}

In the previous approach, the minimum and maximum values of the distribution are
fixed when calculating the scale. However, changing these values might help in improving
quantization. Changing minimum and maximum may introduce
new levels, and change mapping centers thus might allow better approximation for tail values which is crucial.
In the second approach, we modify the technique used by \cite{li2021brecq}, to be robust enough for low and high bit widths. The approach used in \cite{li2021brecq} improves results for asymmetric quantization. However, it fails for symmetric quantization. Therefore, We propose modifications in order to work with asymmetric quantization and symmetric quantization.
In the asymmetric quantization case, the idea is based on changing the minimum and maximum values of the distribution, and then calculating the scale and choosing the best values that minimize the Lp norm of weights. Since optimal p values in the Lp norm are unknown, we try to find the best scale for different values in p space. 
In the case of symmetric quantization, the minimum and maximum adjustment calculated is based on the weights distribution which lies entirely in the positive space or negative space as shown in figure\ref{per-output--channel-weight-dirstibution-mobilenet-v2}. In case of the weight lies entirely in positive space, the maximum of the distribution is fixed and a search is carried out to find the best minimum values that minimize the Lp norm of weights to be used in calculating the scale, the search interval lies between zero and original minimum value in the distribution. In case the weights lie entirely in the negative space, the minimum of the distribution is fixed and a search is carried out to find the best maximum values that minimize the Lp norm of weights to be used in calculating the scale, the search interval lies between zero and original maximum values in the distribution.

Finally, we integrate both approaches min-max adjustment and scale search to push the limits of model performance. For each candidate's minimum and maximum, we search for the best scales that minimize the Lp norm of weights.

\subsection{Variable length encoding}
\label{vlc}
Most of the model's weights distribution is bell-shaped/ logarithmic shape, we take advantage of such distribution and use a variable-length encoding technique to compress weights, thus decreasing the model's memory footprint further. Variable-length encoding leverages from bell shape/ normal distribution, as frequent weights will have a small representation, while infrequent weight values will have a large representation. Therefore by calculating the total bit-rate needed it will be much smaller than the original bit-width (bin number) of the quantized weights. Such techniques could allow mixed-precision quantization, or increase the number of bits for specific layers, while still having a low bit-rate. This could result in a low memory footprint while having a good quantized model. One of the variable encoding techniques that takes advantage of such distribution is Golomb coding. A Golomb-Rice coding \footnote[1]{Code
available at \url{https://github.com/Ritwik00025/Data-Compression/}}  is a variant of Golomb coding which is more hardware-friendly by relying on a divisor which is a power-of-two. This allows the operations to be done through shifting and masking in place of integer division and modulo. Golomb coding relies on encoding a given integer $x$ through computing $q$, the quotient, and $r$, the remainder, according to the two relations:
\begin{equation}
q=\lfloor{\frac{x}{M}}\rfloor,\;r=x-qM
\end{equation}

where M is a tunable parameter. Our experiments showed that M being a power-of-two was the best to achieve the lowest bit-rate. It is also preferable to have such properties to simplify the mathematical operations of the encoding and decoding process. To encode $q$ and $r$, unitary and truncated binary encodings are used respectively. We modify Golomb coding slightly by encoding the case $x=0$ with a special one bit "0" and encode non-zero values for $x$ in the normal way using $(q,r)$ but preceded with "1". This modification results in a better bit-rate compared to the original Golomb coding because of the large number of zero values of the weights. It also simplifies the decoding process by allowing to skip any stream of zeroes of weights (and their corresponding multiplicands) as they don't contribute to the summation. This can help to reduce the latency of the model. The pseudo-code of the whole algorithm is presented in Algorithm \ref{alg:golombcode}.  \\

\begin{algorithm}[h]
\caption{Variable Length Coding using Golomb-Rice Coding}
\label{alg:golombcode}
\small
\begin{algorithmic}
\State{\textbf{Input:} Weights of tensor (W)}
\State{\textbf{Output:} Encoded weights}
\State{\underline {\textbf{Converting Negative values}}}
\For{\texttt{$w_i$ in W}} {
    \If{$w_i < 0$}
    \State \texttt{ $w_i$ = $2* (- w_i) + 1 $}
    \EndIf
}
\EndFor
\State{\underline {\textbf{Compressing Values}}}
\While{$m \leq 2^b $}{
\For{\textbf{$w_i$ in W}} 
    \If{$w_i == 0$}
    \State \texttt{ $en\_w_i$ = "0"}
    \Else
    \State{\texttt{$q = unitary\_enc(floor(w_i/m))$}}
    \State{\texttt{$r= truncated\_binary\_enc(w_i \% m, m)$}}
    \State{\texttt{$en\_w_i=$ "1" $+q+r$}}
    \EndIf
\EndFor

\State{\texttt{$current\_bit\_width=average\_length(en\_w)$}}

\If{$current\_bit\_width < best\_bit\_width$}
 {
 \State{\texttt{$best\_bit\_width = current\_bit\_width$}}
 \State{\texttt{$best\_encoded\_weights = en\_w$}}
 \State{\texttt{$best\_m = m$}}
\EndIf
 }

}
\EndWhile

\State{\underline {\textbf{Decompressing Weights}}}
\For{\textbf{$en\_w_i$ in $en\_w$}} {
\State{\texttt{$i_0=find\_index\_zero(en\_w_i)$}}
\State{\texttt{$inv\_q=inv\_unitary\_enc(en\_w_i[begin:i_0])$}}
\State{\texttt{$inv\_r=inv\_truncated\_binary\_enc(en\_w_i[i_0:end])$}}
\State{\texttt{$w_i= inv\_q*M+inv\_r$}}

\EndFor
}

\end{algorithmic}

\end{algorithm}

VLCQ leverages weight's sparsity and the bell-shape of distribution. Allowing mixed precision and non-mixed precision with a low bit-rate and higher accuracy. This can be achieved by 2-scales quantization and Golomb code encoding. 2-scales allow better approximation for weight values for a different number of levels. While Golomb code encoding accounts for compressing values in order to have a better bit-rate. Hence, low memory footprint. Therefore we could go higher in the number of levels without exceeding the total allowed quota of the required bit-width b. It could also allow us to have a low bit-rate: of up to 2/3 bits while having a high number of levels (more than 8 bins) in each layer.

\begin{figure}[t]
\centering
\includegraphics[width=0.4\textwidth]{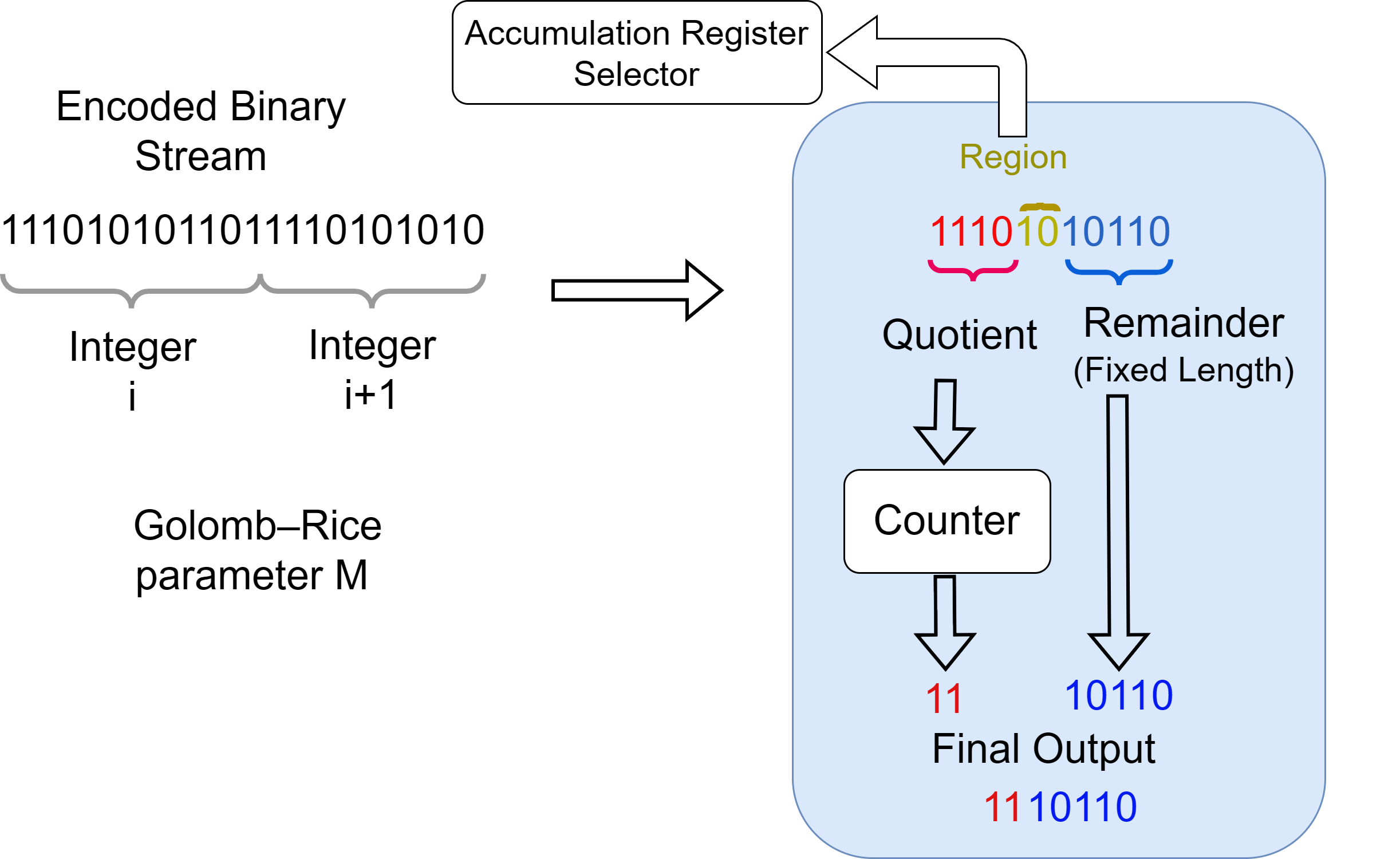}
\caption{Golomb Code decoder design. }
\label{accel_hw_design_decoder}
\end{figure}

\begin{figure}[t]
\centering
\includegraphics[width=0.4\textwidth]{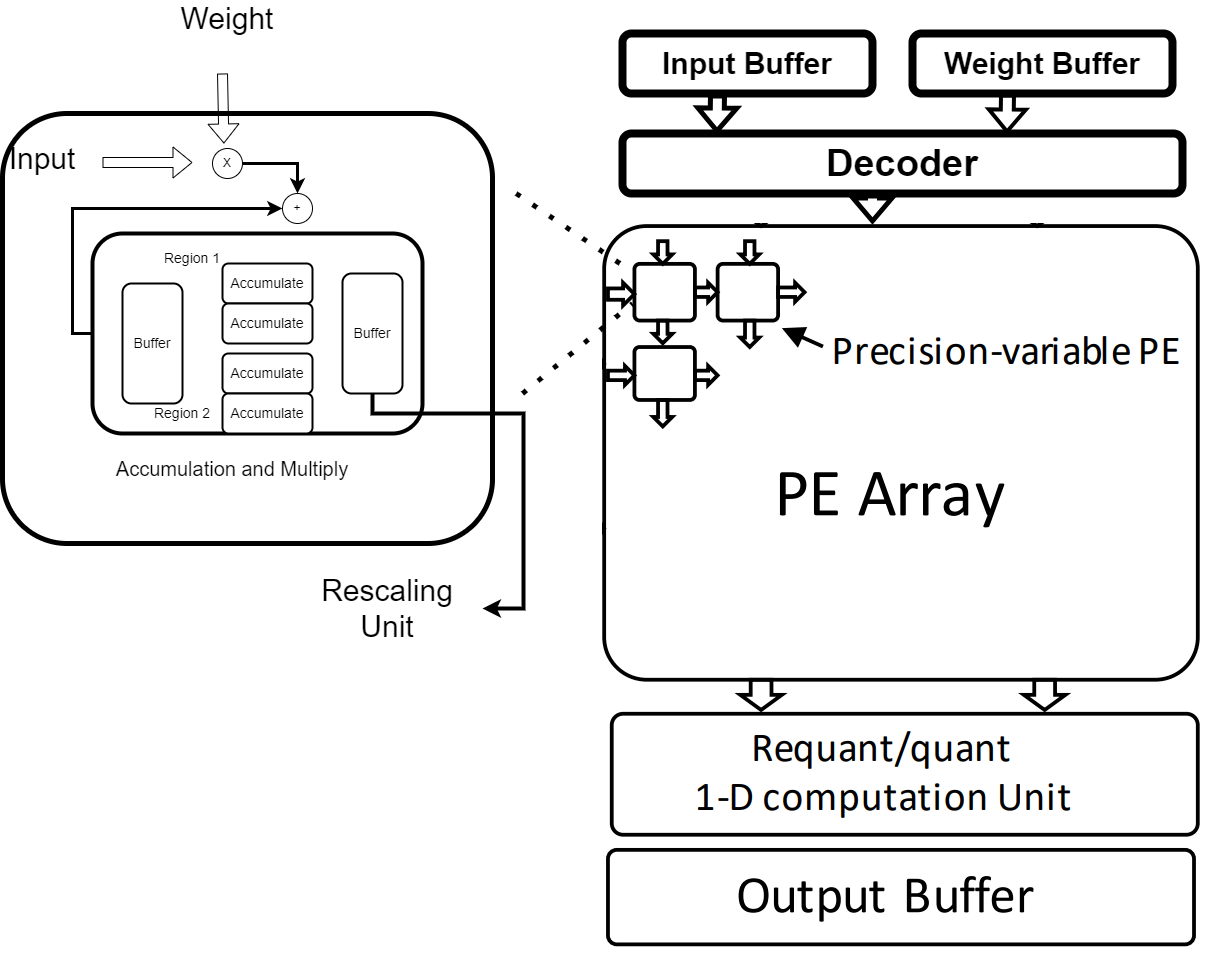}
\caption{Accelerator design to incorporate variable-length encoding. }
\label{accel_hw_design}
\end{figure}

\subsection{Hardware support}
Fixed-length coding was the target of previous research in order not to disrupt the hardware and utilize current processors and GPUs as is. However, this is not a constraint to limit the quantization techniques for deep learning models since it is a common practice to design special hardware accelerators specifically for these models and it can handle any change as long as it enhances both the performance and model compression \cite{HWAccel}.
Therefore, utilizing variable length coding in the custom hardware defies the argument against its practicality. It can be added as an extra early stage in the pipeline of the multiply-accumulate operations so that the effect on the throughput should not be hurt.

\begin{figure}[t]
\centering
\includegraphics[width=0.3\textwidth]{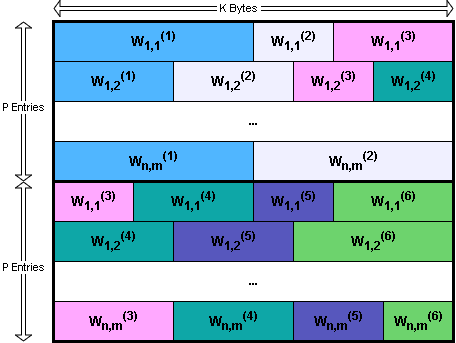}
\caption{Memory Layout to speed up the decoding process. }
\label{memory_layout}
\end{figure}

Two changes are required to implement VLCQ in hardware:
\begin{itemize}
  \item A decoding step to convert variable length coding to fixed one to execute the multiply-accumulate operations, as in figure \ref{accel_hw_design_decoder}. The Golomb code parameter $M$ is a power of two therefore, the remainder is a fixed length value, and no need for conversion to take place as it represents the remainder of the number.

  \item Have multiple accumulators (2 or 4) instead of one, since using different scale values requires accumulating each region separately before multiplying by the scale to dequantize back. This need for multiple accumulators is not limited to our paper. The PWLQ method requires 3 accumulators for the same purpose, as in figure \ref{accel_hw_design}.

  \item The hardware implementation can hide the time penalty since the decoding step will be one stage of the various pipeline stages in the system. The decoder architecture is simple to implement as shown in the below figure.

  \item Since heavy parallelization is needed for model inferences. The decoding of weights need to run in parallel as well. However, if the weights are stored serially in order. This layout must run serially for decoding as well since the beginning of the next weight is not determined until the previous is decoded. The solution for this bottleneck is to lay the weights in a different way as shown in figure \ref{memory_layout}. A block of weights consists of P entries, each entry has K bytes (K = 32 or 64 bits for example). P weights can be decoded in parallel at a time. Initially, the first batch of weights align exactly at the beginning of the first block which can be decoded. Once this batch is decoded to be used for inference, a set of P pointers are needed to point to the beginning of the next batch of weights inside the P entries. Note that a weight can overlap over two consecutive blocks.

  \item Software implementation is another option but would incur a time penalty for decoding the weights. So, it can be used for experimentation or to deploy on devices if the limitation is in memory but not computation. Models are typically huge and easily hit memory limits. So, this is still a viable option.

\end{itemize}

\section{Results}
All Experiments are conducted on a variety of modern deep learning architectures, including ResNet, and MobileNet architectures, using  ImageNet classification tasks.  In all experiments, batch normalization folding is applied before quantization, and bias correction is applied after quantization.  For activations, their values are unquantized and kept in full precision. However, if we quantize the activation into 8 bits with one scale uniform quantization, the results decrease by a constant magnitude of 0.01\%, since we don't use calibration data to adjust activation quantization. During inference, we either apply per-channel quantization or per-tensor quantization. All Experiments are performed using PyTorch.
\label{sec:Results}
\subsection{Parameter setup}
In all experiments, batch normalization folding is applied before quantization, and bias correction is applied after quantization.  For activations, their values are unquantized and kept in full precision. However, if we quantize the activation into 8 bits with one scale uniform quantization, the results decrease by a constant magnitude of 0.01\%, since we don't use calibration data to adjust activation quantization. All Experiments are performed using PyTorch. 
For scale selection with per-channel quantization, we limit search space to 20 different values between $b+1$ and $b-1$, where $b$ is the desired bit-rate. In per-tensor scale selection, the search space is around 300 values. For both cases, we set the $P$ value in lp-norm to $2$ (MSE). In mixed precision quantization, the search space is from 8 downto 3 bit rate (256 levels down to 8 levels).\\
Hardware setup: GPU: 1 x Tesla P100, having 3584 CUDA cores, 16GB GDDR6 VRAM, CPU: single core hyper threaded i.e (1 core, 2 threads) Xeon Processor @ 2.2Ghz, 56MB Cache, RAM: 15.26 GB available.
\subsection{Ablation Study on ImageNet}
In this section, an ablation study is performed using the ImageNet dataset, to investigate the effectiveness of our proposed method VLCQ, and other techniques such as scale selection on other various techniques. We evaluate the top-1 accuracy performance on the test dataset for four popular network architectures: ResNet18, ResNet50, and MobilNetV2. We use torchvision and its pre-trained models for our experiments.

In the previous sections, we introduced some methods independently of the other
methods. In this section, we consider their efficient integration in terms of accuracy, and bit-rate.

\subsubsection{VLCQ}

\begin{figure*}[h!]
    \centering
    \begin{subfigure}{0.3\textwidth}
    \includegraphics[width=\textwidth,keepaspectratio]{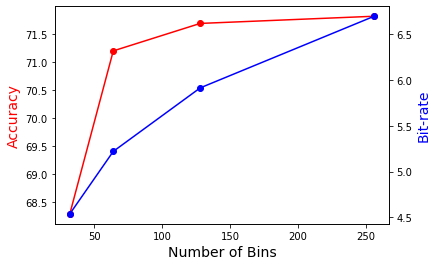}
    \caption{Relationship between bit-rate and number of bins vs top-1 accuracy for per-channel quantization for MobileNetV2. }
    \end{subfigure}
    \begin{subfigure}{0.3\textwidth}
    \includegraphics[width=\textwidth,keepaspectratio]{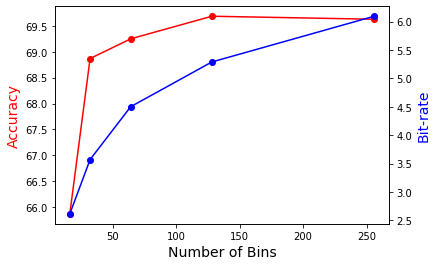}
    \caption{Relationship between bit-rate and number of bins vs top-1 accuracy for per-channel quantization for ResNet18. }
    \end{subfigure}
     \begin{subfigure}{0.3\textwidth}
    \includegraphics[width=\textwidth,keepaspectratio]{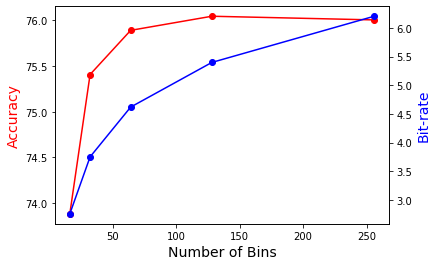}
    \caption{Relationship between bit-rate and number of bins vs top-1 accuracy for per-channel quantization for ResNet50. }
    \end{subfigure}
\caption{Relationship between bit-rate and number of bins vs top-1 accuracy for per-channel quantization. The Red Line refers to the relation between Accuracy and the number of bins and the blue line refers to the relationship between bit-rate and the number of bins.}
\label{per-output--channel-bins-vs-bit-rate}
\end{figure*}

Figure \ref{per-output--channel-bins-vs-bit-rate}, shows the performance of MobileNetV2, ResNet50, and ResNet18 on different numbers of bins and their respective bit-with/rate. It can be seen that VLCQ could reach low-bit-width 3,2 sub-bits for ResNet18 and ResNet50 while having good accuracy.  

\subsubsection{Scale-Selection} \hspace{0pt} \\
For VLCQ, a scale search strategy, Min-Max adjustment, and a combination of both are used to improve overall performance. The scale search strategy is used in both per-tensor quantization and per-channel quantization. For per-channel quantization, the search space is small around 20 different values to search within. While in per-tensor quantization, due to different weights distributions for each channel in the same tensor, the search space is much higher around 300 values. For Min-Max adjustment it's used for per-channel quantization.

\begin{table*}[h!]
\caption{Top-1 accuracy (\%) for various methods. Comparison of accuracy on weights quantized post-training models, \textbf{per-channel methods}. Activations in this case are unquantized and stored with full precision.}
\centering
\footnotesize

\begin{tabular*}{\textwidth}{@{\extracolsep{\fill}}>{\centering}m{0.15\textwidth}>{\centering}m{0.09\textwidth}>{\centering}m{0.07\textwidth}>{\centering}m{0.1\textwidth}>{\centering}m{0.06\textwidth}>{\centering}m{0.1\textwidth}>{\centering}m{0.07\textwidth}>{\centering}m{0.11\textwidth}}
\toprule[1.5pt]
& 
& \multicolumn{2}{c}{MobileNetV2} & \multicolumn{2}{c}{ResNet18} & \multicolumn{2}{c}{ResNet50}\tabularnewline
\cmidrule{3-8}
\multicolumn{1}{p{0.15\textwidth}}{Method} & \multicolumn{1}{p{0.09\textwidth}}{Number of bins}  & \multicolumn{1}{p{0.07\textwidth}}{Bit-rate (W)} & \multicolumn{1}{p{0.1\textwidth}}{Accuracy} & \multicolumn{1}{p{0.06\textwidth}}{Bit-rate (W)} & \multicolumn{1}{p{0.1\textwidth}}{Accuracy} & \multicolumn{1}{p{0.07\textwidth}}{Bit-rate (W)} & \multicolumn{1}{p{0.11\textwidth}}
{Accuracy}\tabularnewline
\cline{3-8}
& & FP32 &  71.880\% & FP32 & 69.758\% & FP32 & 76.150\%\tabularnewline

\toprule[1.5pt]
\multirow{4}{0.15\textwidth}{\centering{}PWLQ} & 512 & 9 & 71.730\% & 9 & 69.622\% & 9 & 76.100\%\tabularnewline
 & 128 & 7 & 71.580\% & 7 & 69.602\% & 7 & 76.080\%\tabularnewline
 & 32 & 5 & 69.220\% & 5 & 69.200\% & 5 & 75.620\%\tabularnewline
 & 16 & 4 & 55.170\% & 4 & 67.076\% & 4 & 73.840\%\tabularnewline
\midrule
\multirow{4}{0.15\textwidth}{\centering{}ACIQ} &  & & & FP32 & 69.700\% & & \tabularnewline
\cline{5-6}
& 256 & - & - & 8 & 68.000\% & 8 & 73.600\%\tabularnewline 
 & 16 & - & - & 4 & 68.300\% & 6 & 75.300\%\tabularnewline
\midrule

\multirow{5}{0.15\textwidth}{\centering{}VLCQ} & 256 & \textbf{6.663} & \textbf{71.876\%} & 6.094 & 69.626\% & 6.203 & 76.002\%\tabularnewline
 & 128 & 5.854 & 71.732\% & 5.288 & 69.684\% & 5.397 & 76.042\%\tabularnewline
 & 64 & 5.105 & 71.368\% & 4.500 & 69.248\% & 4.618 & 75.888\%\tabularnewline
 & 32 & 4.316 & 69.72\% & 3.569 & 68.868\% & 3.749 & 75.406\%\tabularnewline
 & 16 & 3.232 & 60.418\% & 2.611 & 65.872\% & 2.753 & 73.884\%\tabularnewline
\midrule

\multirow{3}{0.15\textwidth}{\centering{}VLCQ MP} & 256 & 5.854 & 71.674\% & - & - & - & -\tabularnewline
 & 256 & 4.468 & 70.616\% & 3.670 & 69.076\% & 3.860 & 75.610\%\tabularnewline
 & 256 & 3.554 & 63.278\% & 2.716 & 66.206\% & 2.898 & 74.168\%\tabularnewline
\midrule
\multirow{6}{0.15\textwidth}{\centering{}VLCQ SS} & 256 & 6.661 & 71.862\% & 6.105 & 69.640\% & 6.210 & 75.994\%\tabularnewline
 & 128 & 5.862 & 71.796\% & 5.325 & 69.592\% & 5.423 & 76.040\%\tabularnewline
 & 64 & 5.149 & 71.330\% & 4.615 & 69.376\% & 4.703 & 76.020\%\tabularnewline
 & 32 & 4.437 & 70.090\% & 3.981 & 68.996\% & 4.049 & 75.516\%\tabularnewline
 & 16 & 3.563 & 64.714\% & 3.157 & 65.404\% & 3.207 & 73.668\%\tabularnewline
\midrule
\multirow{6}{0.15\textwidth}{\centering{}VLCQ Min-Max} & 256 & 6.662 & 71.868\% & 6.094 & 69.728\% & \textbf{6.203} & \textbf{76.120}\%\tabularnewline
 & 128 & 5.854 & 71.802\% & 5.288 & 69.762\% & 5.397 & 76.048\%\tabularnewline
 & 64 & 5.105 & 71.284\% & 4.500 & 69.458\% & 4.618 & 75.968\%\tabularnewline
 & 32 & 4.316 & 69.89\% & 3.569 & 68.846\% & 3.749 & 75.412\%\tabularnewline
 & 16 & 3.233 & 60.468\% & 2.611 & 65.884\% & 2.753 & 73.942\%\tabularnewline
\midrule
\multirow{6}{0.15\textwidth}{\centering{}VLCQ Min-Max SS} & 256 & 6.646 & 71.848\% & 6.104 & 69.738\% & 6.207 & 76.108\%\tabularnewline
 & 128 & 5.851 & 71.772\% & \textbf{5.325} & \textbf{69.790}\% & 5.421 & 76.008\%\tabularnewline
 & 64 & 5.142 & 71.434\% &4.615 & 69.484\% & 4.702 & 76.042\%\tabularnewline
 & 32 & 4.433 & 70.246\% & 3.981 & 69.06\% & 4.049 & 75.616\%\tabularnewline
 & 16 & 3.561 & 65.064\% &3.154 & 65.498\% & 3.205 & 73.604\%\tabularnewline
\bottomrule[1.5pt]
\end{tabular*}

\label{table:comparison-with-competitors}
\end{table*}

\begin{table*}[h]
\caption{Top-1 accuracy (\%) for various methods. Comparison of accuracy on \textbf{per-tensor} quantized weights, post-training. Activations in this case are unquantized and stored with full precision.}

\footnotesize
\begin{tabular*}{\textwidth}{@{\extracolsep{\fill}}>{\centering}m{0.15\textwidth}>{\centering}m{0.09\textwidth}>{\centering}m{0.07\textwidth}>{\centering}m{0.1\textwidth}>{\centering}m{0.06\textwidth}>{\centering}m{0.1\textwidth}>{\centering}m{0.07\textwidth}>{\centering}m{0.11\textwidth}}
\toprule[1.5pt]
& 
& \multicolumn{2}{c}{MobileNetV2} & \multicolumn{2}{c}{ResNet18} & \multicolumn{2}{c}{ResNet50}\tabularnewline
\cmidrule{3-8}
\multicolumn{1}{p{0.15\textwidth}}{\centering{}Method} & \multicolumn{1}{p{0.09\textwidth}}{\centering{}Number of bins} & \multicolumn{1}{p{0.07\textwidth}}{\centering{}Bit-rate (W)} & \multicolumn{1}{p{0.1\textwidth}}{Accuracy} & \multicolumn{1}{p{0.07\textwidth}}{\centering{}Bit-rate (W)} & \multicolumn{1}{p{0.1\textwidth}}{Accuracy} & \multicolumn{1}{p{0.07\textwidth}}{\centering{}Bit-rate (W)} & \multicolumn{1}{p{0.1\textwidth}}{Accuracy}\tabularnewline
\cline{3-8}
& & FP32 &  71.880\% & FP32 & 69.760\% & FP32 & 76.130\%\tabularnewline
\toprule[1.5pt]
\multirow{4}{0.15\textwidth}{\centering{}PWLQ 
} & 512 & 9 & 70.218\% & 9 & 69.622\% & 9 & 76.048\% \tabularnewline
 & 256 & 8 & 70.628\% & 8 & 69.568\% & 8 & 75.932\% \tabularnewline
 & 128 & 7 & 65.700\% & 7 & 69.602\% & 7 & 75.680\% \tabularnewline
 & 32 & 5 & 1.000\% & 5 & 69.200\% & 5 & 69.000\% \tabularnewline
\midrule
\multirow{3}{0.15\textwidth}{\centering{}DFQ} & & FP32 &  71.700\% & FP32 & 69.700\% & & \tabularnewline
\cline{3-6}
& 256 & 8 & 71.200\% & \textbf{8} & \textbf{69.700\%} & - & -\tabularnewline
& 64 & - & - & 6 & 66.300\% & - & -\tabularnewline
\midrule

\multirow{3}{0.15\textwidth}{\centering{}MultiPoint} & & FP32 & 71.780\% & & & &\tabularnewline
\cline{3-4}
& 256 & 8 & 70.700\% & - & - & - & -\tabularnewline
& 16 & - & - & 4 & 61.680\%  & - & -\tabularnewline 
\midrule

\multirow{6}{0.15\textwidth}{\centering{}VLCQ} & 1024 & 7.822 & 71.448\% & 6.319 & 69.640\% & \textbf{6.225} & \textbf{76.062\%}\tabularnewline
 & 512 & 6.680 & 71.020\% & 5.519 & 69.594\% & 5.433 & 75.994\%\tabularnewline
 & 256 & 5.879 & 70.586\% & 4.757 & 69.476\% & 4.669 & 75.816\%\tabularnewline
 & 128 & - & - & 4.003 & 68.658\% & 3.913 & 75.052\%\tabularnewline
 & 64 & - & - & 2.999 & 66.264\% & 2.941 & 68.502\%\tabularnewline
 & 32 & - & - & 2.205 & 34.870\% & 2.146 & 41.292\%\tabularnewline
\midrule
\multirow{4}{0.15\textwidth}{\centering{}VLCQ Non Uniform} & 1024 & \textbf{7.535} & \textbf{71.720\%} & 6.218 & 69.590\% & 6.124 & 75.754\%\tabularnewline
 & 512 & 6.480 & 70.958\% & 5.399 & 69.388\% & 5.322 & 75.954\%\tabularnewline
 & 256 & 5.666 & 69.828\% & 4.603 & 68.596\% & 4.518 & 74.898\%\tabularnewline
 & 128 & 4.909 & 48.190\% & 3.756 & 62.280\% & 3.737 & 69.466\%\tabularnewline
\midrule

\multirow{4}{0.15\textwidth}{\centering{}VLCQ SS} & 256 & 6.142 & 70.058\% & 5.273 & 69.472\% & 5.233 & 75.888\%\tabularnewline
 & 128 & 5.460 & 63.522\% & 4.720 & 69.026\% & 4.738 & 74.960\%\tabularnewline
 & 64 & 4.849 & 45.320\% & 4.210 & 66.036\% & 4.188 & 72.914\%\tabularnewline
 & 32 & - & - & 3.572 & 56.824\% & 3.512 & 66.576\%\tabularnewline
\bottomrule[1.5pt]
\end{tabular*}

\label{table:comparison-with-competitors-per-tesnor}
\end{table*}

Tables \ref{table:comparison-with-competitors} and \ref{table:comparison-with-competitors-per-tesnor} show the effect of adjusting the scale for per-output channel quantization, which is denoted VLCQ SS,  the effect of adjusting the minimum and maximum of the distribution, which is denoted by VLCQ Min-Max and their combination denoted by VLCQ Min-Max SS. VLCQ SS improves accuracy, especially for low bit-rate, e.g. up to 4\% accuracy increase while having the same bit-width and number of levels compared to normal 2 scales binning and no binning modes. However, VLCQ Min-Max improves accuracy especially for ResNet18 and ResNet50, for high bit-rate, allowing us to reach near-exact original floating-point accuracy with low bit-rate. For VLCQ Min-Max SS near-exact original floating-point accuracy is maintained up to 32 bins quantization.
\subsubsection{One Scale uniform quantization}
We test different schemes of scale selection: 1-Exploring search space of possible scales, 2-Finding best scales by changing minimum and maximum of the weights,  on one scale uniform quantization. Experiments are conducted using the ImageNet dataset on ResNet18, ResNet50, and MobileNetV2 models. Bias correction is applied to all layers. 




\subsubsection{Mixed Precision}
Different models are evaluated based on the discussed mixed-precision strategy, using the ImageNet dataset. 

It can be seen from tables \ref{table:comparison-with-competitors}, \ref{table:comparison-with-competitors-per-tesnor} and \ref{table:mixedpercision} the potential of mixed precision and its ability to push the limit of PTQ further. Tables \ref{table:comparison-with-competitors} and \ref{table:mixedpercision}, show the effect of using different bit-width on each region, in the proposed quantization strategy. Mixed precision frequently beats unified precision, especially when utilizing extremely low-bit,  with up to 3\% accuracy increase with the same bit width.

\npdecimalsign{.}
\nprounddigits{3}
\begin{table*}[t]
\caption{Top-1 accuracy (\%) for MobileNetV2, ResNet18, and ResNet50 when using \textbf{per-region bit-allocation mixed precision strategy, per-region per-layer bit-allocation mixed precision strategy, and per-region bit-allocation mixed precision strategy with scale selection.} Model weights are quantized, but activations here are unquantized and kept in full precision.}
\footnotesize
\begin{tabular*}{1.03\textwidth}{@{\extracolsep{\fill}}>{\centering}m{0.12\textwidth}>{\centering}m{0.08\textwidth}>{\centering}m{0.06\textwidth}>{\centering}m{0.06\textwidth}>{\centering}m{0.1\textwidth}>{\centering}m{0.06\textwidth}>{\centering}m{0.1\textwidth}>{\centering}m{0.06\textwidth}>{\centering}m{0.1\textwidth}}
\toprule[1.5pt]
&
&
&
\multicolumn{2}{c}{MobileNetV2} & \multicolumn{2}{c}{ResNet18} & \multicolumn{2}{c}{ResNet50}\tabularnewline
\cmidrule{4-9}
\multicolumn{1}{p{0.12\textwidth}}{Method} & \multicolumn{1}{p{0.08\textwidth}}{Number of bins} & \multicolumn{1}{p{0.06\textwidth}}{Target bit-rate} & \multicolumn{1}{p{0.06\textwidth}}{Bit-rate (W)} & \multicolumn{1}{p{0.1\textwidth}}{Accuracy FP32: 71.880\%} & \multicolumn{1}{p{0.06\textwidth}}{Bit-rate (W)} & \multicolumn{1}{p{0.1\textwidth}}{Accuracy FP32: 69.760\%} & \multicolumn{1}{p{0.06\textwidth}}{Bit-rate (W)} & \multicolumn{1}{p{0.1\textwidth}}{Accuracy FP32: 76.130\%}\tabularnewline
\toprule[1.5pt]
\multirow{3}{0.12\textwidth}{VLCQ MP} & 256 & 6 & 5.854 & 71.674\% & 4.533 & 69.632\% & 4.651 & 76.008\%\tabularnewline
 & 256 & 5 & 4.468 & 70.616\% & 3.670 & 69.076\% & 3.860 & 75.610\%\tabularnewline
 & 256 & 4 & 3.554 & 63.278\% & 2.716 & 66.206\% & 2.898 & 74.168\%\tabularnewline
\midrule
\multirow{5}{0.12\textwidth}{VLCQ MP SS} & 256 & 6 & 5.885 & 71.784\% & 5.325 & 69.618\% & 5.423 & 76.020\%\tabularnewline
 & 256 & 5 & 5.193 & 71.348\% & 4.615 & 69.378\% & 4.704 & 76.010\%\tabularnewline
 & 256 & 4 & 4.466 & 70.068\% & 3.981 & 68.990\% & 4.049 & 75.570\%\tabularnewline
 & 256 & 3 & 3.587 & 65.102\% & 3.157 & 65.422\% & 3.207 & 73.712\%\tabularnewline
 & 256 & 2 & 2.799 & 10.106\% & 2.566 & 39.47\% & 2.597 & 59.532\%\tabularnewline
\midrule
\multirow{4}{0.12\textwidth}{VLCQ MP (per layer - region)} & 256 & 6 & 5.140 & 71.224\% & 4.520 & 69.424\% & 4.653 & 75.948\%\tabularnewline
 & 256 & 5 & 4.382 & 69.014\% & 3.635 & 69.208\% & 3.849 & 75.654\%\tabularnewline
 & 256 & 4 & 3.359 & 62.518\% & 2.736 & 66.862\% & 2.871 & 74.704\%\tabularnewline
 & 256 & 3 & 3.339 & 61.358\% & 2.716 & 66.260\% & 2.837 & 73.832\%\tabularnewline
\bottomrule[1.5pt]
\end{tabular*}
\label{table:mixedpercision}
\end{table*}
\npnoround

\subsection{Comparison with other techniques}
In this section, we provide a comprehensive comparison between VLCQ and other quantization systems including PWLQ, DFQ, ACIQ, AdaRound, and AdaQuant, in various tasks as segmentation, classification, and object detection . Note that activations here are unquantized and kept in full precision. For VLCQ bias correction is used. Several methods which are discussed in the above subsection are integrated with VLCQ as scale selection, Min-Max adjustment, mixed-precision, and their combination. The LP value is set to 2. In addition, we demonstrate that VLCQ with its variants could be integrated with other methods achieving high performance.

\subsubsection{Image Classification}
We provide a comprehensive comparison between VLCQ and other quantization systems including PWLQ, DFQ, ACIQ, AdaRound, and AdaQuant, using Imagenet dataset, for classification tasks.
In table \ref{table:comparison-with-competitors}, weights are quantized, while activations are kept at full precision.
However, if we quantize the activation into 8 bits with one scale uniform quantization, the results decrease by a constant magnitude of 0.01\%, since we don't use calibration data to adjust activation quantization. We re-run most of the state-of-the-art papers based on their open-source implementation, without having activation quantized and their results match the reported results since they use calibration data for activation quantization. Therefore keeping activation without quantization or with 8-bit quantization has no effect on the results.
VLCQ outperforms other methods while having a low bit-rate. In addition to that, VLCQ could quantize ResNet18, below 4 bit-width without access to data. 
PWLQ works by dividing distribution into 3 dense regions and two sparse regions for the dense region uniform quantization with b-bit width is applied while for both sparse regions uniform asymmetric quantization with b-1 bit-width is applied. That results in allocating $2* 2^B$ levels for each layer and an extra bit to determine which region it belongs to. Thus the total number of levels in each layer is $2* 2^B$ and bit-rate is $B+1$. 
VLCQ outperform PWLQ using  256, 64, and 16 bins for the MobileNetV2 model.

For ResNet50 and ResNet18, VLCQ achieves good performance (3-4\% accuracy loss) with sub 3 bit-width making it the first data-free PTQ method to achieve good performance with small bit-width.
For 256 bins there is a margin of 0.02\% increase with lower bit-width, VLCQ Min-Max, and  VLCQ Min-Max SS could achieve a near-perfect accuracy of 76.108\% and 76.120\% with 7 sub-bit-width while PWLQ achieves 76.100\% with full 9 bits, with 512 bins.

VLCQ  achieves similar results compared to PWLQ but with smaller bit-rates. For ResNet18, PWLQ achieves 69.622\% accuracy while VLCQ Min-Max SS achieves 69.790\% accuracy with 5.3 bits which is better than the FP32 accuracy. The reason for why quantization can accidentally achieve better results, is because when the weights get quantized, the new values can luckily achieve better accuracy. 
Furthermore, VLCQ MP achieves around 1\% error loss, which is similar to BRECQ (BRECQ uses calibration dataset to find optimal rounding strategy)  which achieves 1\% error loss on MobileNetV2. However, VLCQ SS achieves around 1.6\% error with a sub 5 bit-rate in MobileNetV2.  Thus, VLCQ is the first DFQ-PTQ technique to achieve around 1-2\% error with a low bit-rate for MobileNetV2, ResNet50, and ResNet18. 


Table \ref{table:comparison-with-competitor-fintune}, shows top 1\% accuracy for various selected best results in the proposed different VLCQ variants (which do not use calibration data) with other PTQ methods that use calibration data. VLCQ clearly matches or supersedes the performance of AdaQuant. Compared to AdaRound, VLCQ is tightly close to its performance although VLCQ is data-free.

\begin{table*}[h!]
\caption{Top-1 accuracy (\%) for VLCQ against various PTQ methods with calibration data. Comparison of accuracy on \textbf{per-channel} quantized weights, post-training. Activations in this case are unquantized and stored with full precision.}

\footnotesize
\begin{tabular*}{\textwidth}{@{\extracolsep{\fill}}>{\centering}m{0.15\textwidth}>{\centering}m{0.09\textwidth}>{\centering}m{0.07\textwidth}>{\centering}m{0.1\textwidth}>{\centering}m{0.06\textwidth}>{\centering}m{0.1\textwidth}>{\centering}m{0.07\textwidth}>{\centering}m{0.11\textwidth}}
\toprule[1.5pt]
& 
& \multicolumn{2}{c}{MobileNetV2} & \multicolumn{2}{c}{ResNet18} & \multicolumn{2}{c}{ResNet50}\tabularnewline
\cmidrule{3-8}
\multicolumn{1}{p{0.15\textwidth}}{\centering{}Method} & \multicolumn{1}{p{0.09\textwidth}}{\centering{}Number of bins} & \multicolumn{1}{p{0.07\textwidth}}{\centering{}Bit-rate (W)} & \multicolumn{1}{p{0.1\textwidth}}{Accuracy} & \multicolumn{1}{p{0.07\textwidth}}{\centering{}Bit-rate (W)} & \multicolumn{1}{p{0.1\textwidth}}{Accuracy } & \multicolumn{1}{p{0.07\textwidth}}{\centering{}Bit-rate (W)} & \multicolumn{1}{p{0.1\textwidth}}{Accuracy}\tabularnewline
\toprule[1.5pt]
\multirow{2}{0.15\textwidth}{\centering{}AdaRound} & & FP32 & 71.720\% &  & 69.680\% &  & 76.070\%\tabularnewline
\cmidrule{3-8}
 & 16 & 4 & 69.780\% & 4 & 69.710\% & 4 & 75.230\% \tabularnewline
\midrule
\multirow{2}{0.15\textwidth}{\centering{}AdaQuant} &  & FP32 & 72.490\% &  & 71.080\% &  & 
77.000\%\tabularnewline
\cmidrule{3-8}
 & 16 & 4 & 44.780\% & 4 & 68.820\% & 4 & 75.220\% \tabularnewline
\midrule
\multirow{3}{0.15\textwidth}{\centering{}VLCQ} &  & FP32 & 71.880\% & & 69.760\%& & 76.130\% \tabularnewline
\cmidrule{3-8}
 & 32 & 4.433 & 70.246\% & 3.981 & 69.060\%& 3.749& 75.412\% \tabularnewline
& 16 & 3.561 & 65.064\% & 2.611 & 65.872\%& 2.753& 73.940\% \tabularnewline
\bottomrule[1.5pt]
\end{tabular*}

\label{table:comparison-with-competitor-fintune}
\end{table*}

Table \ref{table:comparison-with-competitors-per-tesnor}, shows the top 1\% accuracy for different per-tensor quantization techniques, This is cheap to implement on any existing low-precision integer hardware. VLCQ outperforms most of the existing methods, showing acceptable performance for 8,6,7 bit-rate quantization.

Table \ref{table:memory_flops} shows the model size and the FLOPs needed. For the memory footprint, VLCQ occupies less memory than PWLQ and uniform quantization for two reasons. First reason is that Golomb code encoding is applied which compresses weights representation. Secondly, PWLQ and uniform quantization apply asymmetric affine quantizers which need to store offset FP32 values for each channel. Unlike VLCQ which uses symmetric quantizers. In terms of FLOPs, VLCQ has a similar number of FLOPS compared to uniform quantizers, as in VLCQ after the MAC is applied in each region its descaled then the addition operation takes place in FP space to get the final activation where this operation depends on the tensor size. In uniform quantizers, since they use affine asymmetric  quantizers after the MAC operation takes place and tensor is dequantized to original space, the offset of the quantizers is added to the dequantized weights, therefore this operation depends on size of tensor. Hence, VLCQ does not suffer from an increase in FLOPs compared to normal methods. Unlike PWLQ which divides space into three regions, three additions need to be performed in FP space to construct output activation, in addition to 2 FP addition to add offset, since the quantizer operator is affine quantizer.
\begin{figure*}[t!]
\centering
\begin{subfigure}[b]{0.3\textwidth}
    \includegraphics[width=\textwidth]{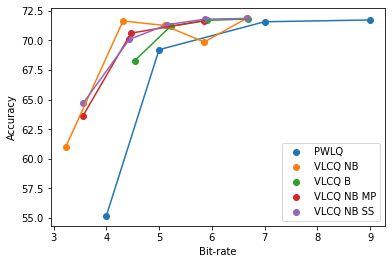} 
    \caption{MobileNetV2 model}
\end{subfigure}
 \begin{subfigure}[b]{0.3\textwidth}
    \includegraphics[width=\textwidth]{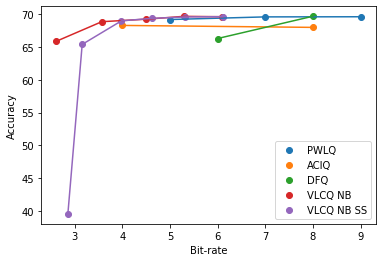}
    \caption{ResNet18 model}
\end{subfigure}
 \begin{subfigure}[b]{0.3\textwidth}
    \includegraphics[width=\textwidth]{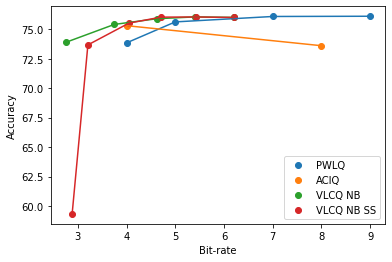}
    \caption{ResNet50 model}
\end{subfigure}
\caption{Top-1 accuracy (\%) and bit-widths for the indicated model. Comparison of accuracy on weights quantized post-training models,
per-channel methods. Activations in this case are unquantized and stored with full precision.}
\label{model_bitrate_accuracy_visualization}
\end{figure*}
\begin{table}[h]
\caption{The memory size of the model and FLOPs for \textbf{MobileNetV2} for per-channel quantization.}
\centering
\footnotesize
\begin{tabular}{p{4em}>{\centering}p{3em}>{\centering}p{4em}>{\centering}p{4em}>{\centering}p{5em}}\toprule[1.5pt]
Technique                             & Bit-rate & Number of Bins & Model Size (13.5MB)  & FLOPs                       \tabularnewline\toprule[1.5pt]
\multirow{3}{*}{PWLQ}                 & 8        & 256            & 3.5726          & \multirow{3}{*}{26,932,832} \tabularnewline
                                      & 7        & 128            & 3.136           &                             \tabularnewline
                                      & 6        & 64             & 2.7             &                             \tabularnewline \midrule
\multirow{3}{4em}{\parbox{4em}{Uniform Quant.}} & 8        & 256            & 3.51755         & \multirow{3}{*}{13,574,608} \tabularnewline
                                      & 7        & 128            & 3.0815          &                             \tabularnewline
                                      & 6        & 64             & 2.645           &                             \tabularnewline\midrule
\multirow{3}{*}{VLCQ}                 & 6.681    & 256            & 2.97            & \multirow{3}{*}{13,574,608} \tabularnewline
                                      & 5.854    & 128            & 2.609           &                             \tabularnewline
                                      & 5.105    & 64             & 2.282           &                             \tabularnewline\bottomrule[1.5pt]
\end{tabular}
\label{table:memory_flops} 
\end{table}
Figure \ref{model_bitrate_accuracy_visualization}
shows the relationship between bit-width and accuracy for different techniques for the different models. It can be seen that VLCQ achieves the same accuracy or outperforms most of the other techniques while requiring fewer bits.
Golomb coding achieves low-entropy for cases where the data is exponentially-distributed. It turns out that this case occurs in one-scale uniform quantization and two-scale quantization with different modes. However, Golomb coding performance degrades with other quantization techniques such as PWLQ, BRECQ, and DFQ, as shown in table \ref{table:bit-rate-comparison-with-competitors}. This case happens as DFQ, PWLQ, and BRECQ try to allocate as many bins as possible in order to accurately approximate the weights distribution, which results in a flat  weights distribution rather than an exponential one.
As shown in table \ref{table:comparison-with-competitors},  using an arbitrary number of bins on each region improves performances, as it allows a better approximation for the region which have more floating-point levels, furthermore, the bit-rate is not affected much, thus allowing us to increase the number of bits without need to worry about bit-rate. Therefore achieving high performance with a small bit-rate.

\begin{table}[h]
\caption{Bit-rate estimation using variable length encoding in different quantization
techniques for \textbf{MobileNetV2}.}
\centering %
\footnotesize
\begin{tabular}{p{5.5em}>{\centering}p{6.5em}>{\centering}p{5em}}\toprule[1.5pt]
Technique  & Number of bins  & Bit-rate \tabularnewline
\toprule[1.5pt]
\multirow{3}{5.5em}{PWLQ} & 512  & 8.464 \tabularnewline
 &  64  & 6.103 \tabularnewline
 &  32  & 5.125 \tabularnewline
\midrule
DFQ  &  256  & 8.527 \tabularnewline
\midrule
BRECQ  & 16  & 4.010 \tabularnewline \bottomrule[1.5pt]
\end{tabular}\label{table:bit-rate-comparison-with-competitors} 
\end{table}

Based on the previously discussed results for per-channel quantization. In high bit-rate quantization (8-7-6 bits) it is better to use VLCQ or VLCQ SS Min-Max as the magnitude of the error is around 2\% or less for most of the models. However, for small bit-rate quantization, VLCQ SS Min-Max is the best choice as the magnitude of the error is around 6\% or less for most of the models.  For per-tensor quantization, it is better to use VLCQ or VLCQ SS.

In terms of the memory footprint required for VLCQ or VLCQ SS, VLCQ MP or VLCQ Min-Max, or VLCQ SS Min-Max, for per-channel quantization, each channel needs to have enough memory to store 2 scales for each region. In per-tenor quantization, each layer needs to store only the 2 scales for each region. Since the regions are divided around zero, therefore there is no need for an extra bit to determine the region. In the case of VLCQ non-uniform 4 scales needed to be stored for per-layer quantization. 

\subsubsection{VLCQ Extendability}
VLCQ can be easily integrated into other techniques such as: BRECQ, resulting in overall improved performance.
Table \ref{table:BRECQ_BRECQ_VLCQ} shows that applying our scale selection method to BRECQ for 4-bit quantization outperforms the original method. However, the original BRECQ method outperforms in 2-bit quantization with respect to BRECQ with the proposed scale selection, which might indicate that the parameter of BRECQ original method scale is adjusted to work well for 2 bits quantization, leaving room to improve other bit-widths. Furthermore, BRECQ with our proposed 2-scales method, outperforms the original BRECQ method for 2 \& 4 bit-width quantization.

\begin{table}[h]

\caption{Top-1 accuracy (\%) for \textbf{MobileNetV2} (FP32: 72.490\%) for BRECQ original model and when integrated to the proposed scale selection, and 2 scales method.}
\centering
\footnotesize
\begin{tabular}{l>{\centering}p{40pt}>{\centering}p{30pt}>{\centering}p{40pt}}
\toprule[1.5pt]
Method  & Accuracy & Bit-width Weights & Bit-width Activation\tabularnewline
\toprule[1.5pt]
\multirow{2}{*}{BRECQ} & 71.420\% & 4 & \multirow{6}{*}{8}\tabularnewline
 &  55.580\% & 2 & \tabularnewline
\cline{1-1} \cline{2-3} 
\multirow{2}{*}{BRECQ SS Min-Max}  & 72.004\% & 4 & \tabularnewline
 &  56.19\% & 2 & \tabularnewline
\cline{1-1} \cline{2-3} 
\multirow{2}{*}{BRECQ 2-Scales} &  72.306\% & 4 & \tabularnewline
 &  65.804\% & 2 & \tabularnewline
\bottomrule[1.5pt]
\end{tabular}
\label{table:BRECQ_BRECQ_VLCQ}
\end{table}

\subsubsection{Other tasks}
To demonstrate the robustness and scalability of our proposed approach, we extend the VLCQ approach to other computer vision tasks such as semantic segmentation on the Pascal VOC segmentation dataset \cite{pascal-voc-2012} and object identification on Pascal VOC object detection challenge \cite{pascal-voc-2007}.
\paragraph{Semantic segmentation}
We use VLCQ and its variants on DeepLab-v3+ with a MobileNetV2 backbone. On the Pascal VOC segmentation task, performance is measured using mean intersection over union (mIoU).  In our experiments, we use the public Pytorch repository implementation to measure performance. Table \ref{table:comparison-with-competitors-per-channel_pascal_voc_ss}  compares the results of the mIoU score of quantized models between PWLQ and VLCQ and its variants. DeepLab-v3+ with MobileNetV2 backbone is quite a challenging model, since it has several layers of weight with very large ranges (e.g., [-54.4, 64.4]). This makes it hard to have a good quantization performance specially for low bit-rate. VLCQ Min-Max SS achieves higher results than PWLQ in 128 and 32 levels by magnitude of 0.44\% and 1.4\%. Adding Min-Max SS, SS, Min-Max to VLCQ compared to normal VLCQ shows promising results. For low bit-rate of 32 levels, VLCQ Min-Max SS improves over VLCQ SS by magnitude of 10\%. For 64 levels quantization, VLCQ Min-Max SS improves results compared to VLCQ SS by magnitude of 11\%. Similarly, VLCQ Min-Max improves results compared to VLCQ SS by magnitude of 5\% in 64 levels quantization. For 64 levels quantization, VLCQ Min-Max SS improves results compared to VLCQ by magnitude of 6\%.

\begin{table}[h!]
\caption{ Comparison of mIOU on Pascal VOC semantic segmentation  dataset.}
\footnotesize
\begin{tabular}{@{\extracolsep{\fill}}c>{\centering}p{50pt}>{\centering}p{30pt}>{\centering}p{30pt}>{\centering}p{30pt}}
\toprule[1.5pt]
 &  &  & \multicolumn{2}{c}{MobileNetV2}\tabularnewline
 \cmidrule{4-5}
Method & Per channel & Number of bins & Bit rate (W) & mIOU\tabularnewline
 \cmidrule{4-5}
 & & & FP32 & 70.75\% \tabularnewline
 \toprule[1.5pt]
\multirow{3}{*}{PWLQ} & \Checkmark{} & 512 & 9  & 70.76\%\tabularnewline
 & \Checkmark{} & 128 & 7  & 70.08\% \tabularnewline
 & \Checkmark{} & 32 & 5  & 56.49\% \tabularnewline
\midrule
\multirow{3}{*}{VLCQ} & \Checkmark{} & 256 & 6.458  & 70.16\%\tabularnewline
 & \Checkmark{} & 128 & 5.6571  & 69.46\%\tabularnewline
 & \Checkmark{} & 64 & 4.911  & 62.00\% \tabularnewline
 \midrule
\multirow{6}{*}{\parbox{50pt}{\centering{}VLCQ SS}} & \Checkmark{} & 1024  & 8.460  & \textbf{70.81\%}\tabularnewline
 & \Checkmark{} & 512  & 7.520  & 70.76\%\tabularnewline
 & \Checkmark{} & 256  & 6.463  & 70.31\%\tabularnewline
 & \Checkmark{} & 128 & 5.669  & 69.46\%\tabularnewline
 & \Checkmark{} & 64 & 4.954  & 58.11\% \tabularnewline
 & \Checkmark{} & 32 & 4.268  & 41.71\% \tabularnewline
\midrule
\multirow{3}{*}{\parbox{50pt}{\centering{}VLCQ Min Max}} & \Checkmark{} & 256  & 5.675  & 70.14\% \tabularnewline
 & \Checkmark{} & 128 & 4.929  & 66.04\%\tabularnewline
 & \Checkmark{} & 64 & 4.156  & 62.30\% \tabularnewline
\midrule
\multirow{4}{*}{\parbox{50pt}{\centering{}VLCQ Min Max SS}} & \Checkmark{} & 256  & 6.451  & 70.58\% \tabularnewline
 & \Checkmark{} & 128 & 5.659  & 70.44\%\tabularnewline
 & \Checkmark{} & 64 & 4.947  & 69.23\% \tabularnewline
 & \Checkmark{} & 32 & 4.262  & 57.04\% \tabularnewline
\bottomrule[1.5 pt]
\multirow{4}{*}{VLCQ} & \XSolidBrush{} & 1024 & 7.436  & 68.87\%\tabularnewline
 & \XSolidBrush{} & 512 & 6.388  & 67.15\%\tabularnewline
 & \XSolidBrush{} & 256 & 5.588  & 63.21\% \tabularnewline
 & \XSolidBrush{} & 128 & 4.840  & 58.98\% \tabularnewline
\midrule
\multirow{2}{*}{DFQ} &  &  & FP32 & 72.45\% \tabularnewline
 \cmidrule{4-5}
 & \XSolidBrush{} & 256 & 8 & \textbf{72.33\%} \tabularnewline
\bottomrule[1.5 pt]
\end{tabular}
\label{table:comparison-with-competitors-per-channel_pascal_voc_ss}
\end{table}


\begin{table}[h!]
\caption{  Comparison of mAP on Pascal VOC object detection  dataset.}
\centering
\footnotesize
\begin{tabular}{@{\extracolsep{\fill}}>{\centering}m{50pt}>{\centering}m{30pt}>{\centering}m{30pt}>{\centering}m{30pt}>{\centering}m{30pt}}
\toprule[1.5pt]
 &  &  & \multicolumn{2}{c}{MobileNetV2}\tabularnewline
\cmidrule{4-5}
Method & Per channel & Number of bins & Bit rate (W) & mAP \tabularnewline
 \cmidrule{4-5}
 & & & FP32 & 68.6\% \tabularnewline
\toprule[1.5pt]
\multirow{4}{*}{\centering{}PWLQ 
} & \checkmark & 512 & 9 &68.51\% \tabularnewline
 & \checkmark & 128& 7 &68.42\% \tabularnewline
 & \checkmark &32 & 5 & 68.32\% \tabularnewline
\midrule
\multirow{4}{*}{\centering{}VLCQ}& \checkmark & 256 & 6.368 & \textbf{68.71\%} \tabularnewline
 & \checkmark & 128 & 5.570 & 68.64\%\tabularnewline
 & \checkmark & 64 &4.815 & 68.08\% \tabularnewline
 & \checkmark & 32 & 4.043& 66.53\% \tabularnewline
 & \checkmark & 16 & 3.054 & 52.59\% \tabularnewline
\midrule
\multirow{4}{*}{\parbox{50pt}{\centering{}VLCQ SS}}& \checkmark  & 256& 6.370 & \textbf{68.67\%} \tabularnewline
 & \checkmark & 128& 5.579 & 68.56\%\tabularnewline
 & \checkmark & 64& 4.833 & 68.58\% \tabularnewline
 & \checkmark & 32& 4.104 & 67.47\% \tabularnewline
& \checkmark & 16& 3.242 & 59.80\% \tabularnewline
\midrule
\multirow{5}{*}{\parbox{50pt}{\centering{}VLCQ Min Max}} & \checkmark & 256& 6.379 & \textbf{68.66\%} \tabularnewline
 & \checkmark & 128& 5.575 & 68.70\%\tabularnewline
 & \checkmark & 64&4.802 &68.52\% \tabularnewline
 & \checkmark & 32& 4.003 & 68.62\% \tabularnewline
 & \checkmark & 16& 2.962& 67.12\% \tabularnewline
 & \checkmark & 8& 1.950 & 39.85\% \tabularnewline
\midrule
\multirow{4}{*}{\parbox{50pt}{\centering{}VLCQ Min Max SS}} & \checkmark& 256 & 6.353 & \textbf{68.60\%} \tabularnewline
 & \checkmark & 128 & 5.557 & 68.69\%\tabularnewline
 & \checkmark & 64& 4.818 &68.58\% \tabularnewline
 & \checkmark & 32& 4.096 & 68.36\% \tabularnewline
 & \checkmark & 16& 3.235 & 67.43\% \tabularnewline
 & \checkmark & 8& 2.514 & 59.41\%\tabularnewline
\bottomrule[1.5 pt]
\multirow{4}{*}{\centering{}VLCQ} & \xmark & 256 & 5.340 & {68.39\%} \tabularnewline
 & \xmark & 128& 4.570 & 66.47\%\tabularnewline
 & \xmark & 64& 3.800&  60.53\% \tabularnewline
\midrule
\multirow{4}{*}{\parbox{50pt}{\centering{}VLCQ SS}} & \xmark & 256& 5.500 & \textbf{68.47\%} \tabularnewline
 & \xmark & 128 & 4.870 & 67.81\%\tabularnewline
 & \xmark & 64 & 4.220 & 65.68\% \tabularnewline
 & \xmark & 32 & 3.620 & 59.51\% \tabularnewline
\midrule
DFQ &\xmark & 256& 8 & {67.91\%} \tabularnewline
\bottomrule[1.5pt]
\end{tabular}
\label{table:comparison-with-competitors-per-channel_pascal_voc_obj_detect}
\end{table}

\begin{table*}[h]

\caption{Comparison of mAP on Pascal VOC object detection dataset for Yolo models.}

\footnotesize
\begin{tabular*}{\textwidth}{@{\extracolsep{\fill}}>{\centering}m{0.1\textwidth}>{\centering}m{0.05\textwidth}>{\centering}m{0.06\textwidth}>{\centering}m{0.1\textwidth}>{\centering}m{0.05\textwidth}>{\centering}m{0.1\textwidth}>{\centering}m{0.05\textwidth}>
{\centering}m{0.1\textwidth}>{\centering}m{0.05\textwidth}>{\centering}m{0.1\textwidth}}
\toprule[1.7pt]
& 
& \multicolumn{2}{c}{YOLOv11n} & \multicolumn{2}{c}{YOLOv8n} & \multicolumn{2}{c}{YOLOv5s} & \multicolumn{2}{c}{YOLOv3}\tabularnewline
\cmidrule{3-10}
\multicolumn{1}{p{0.1\textwidth}}{\centering{}Method} & \multicolumn{1}{p{0.05\textwidth}}{\centering{}Number of bins} & \multicolumn{1}{p{0.06\textwidth}}{\centering{}Bit-rate (W)} & \multicolumn{1}{p{0.1\textwidth}}{mAP} & \multicolumn{1}{p{0.05\textwidth}}{\centering{}Bit-rate (W)} & \multicolumn{1}{p{0.1\textwidth}}{mAP} & 
\multicolumn{1}{p{0.05\textwidth}}{\centering{}Bit-rate (W)} & \multicolumn{1}{p{0.1\textwidth}}{mAP} &
\multicolumn{1}{p{0.05\textwidth}}{\centering{}Bit-rate (W)} & \multicolumn{1}{p{0.1\textwidth}}{mAP}

\tabularnewline
\toprule[1.5pt]
&  & FP32 & 0.548  & FP32 & 0.521 & FP32 & 0.594  & FP32 &  0.684\tabularnewline
\cline{3-10}

\multirow{4}{0.15\textwidth}{\centering{}VLCQ} 
 & 256 & 5.96 & 0.547 & 5.89 & 0.52 & 5.91 & 0.593 & 5.59  & 0.684\tabularnewline
 & 128 & 5.16 & 0.542 &5.09  & 0.518 & 5.11 & 0.592 & 4.82  & 0.684\tabularnewline
& 64 & 4.35 & 0.536 & 4.28 & 0.509  & 4.3& 0.588 & 4  & 0.683 \tabularnewline
 & 32 & 3.52 & 0.516 &  3.4 & 0.486 & 3.4 & 0.577 & 3.016  & 0.679\tabularnewline

\midrule

\multirow{4}{0.15\textwidth}{\centering{}VLCQ \\Min-Max} & 256 & 5.95 & 0.547 & 5.89  & 0.52 & 5.91  &  0.593 &5.59 & 0.684 \tabularnewline
 & 128 & 5.15 & 0.544 & 5 & 0.519  & 5.11 & 0.594 & 4.82  & 0.684\tabularnewline
 & 64 & 4.34 & 0.538 & 4.28 & 0.511 & 4.3 & 0.59 & 4 & 0.683 \tabularnewline
 & 32 & 3.5 & 0.517 & 3.4  & 0.486 & 3.4 & 0.577 & 3.01  & 0.679\tabularnewline
  & 16 & - & - & 2.43 & 0.197 &2.43 & 0.307 & 2.11  & 0.504 \tabularnewline
 
\midrule

\multirow{4}{0.15\textwidth}{\centering{}VLCQ \\Min-Max SS} & 256 & 5.94 & 0.547 & 5.889  &  0.521 & 5.91 & 0.594 & 5.601 & 0.684 \tabularnewline
 & 128 & 5.15 & 0.544 &5.1  &0.517   &  5.12 & 0.593 & 4.88 & 0.684\tabularnewline
 & 64 & 4.38 & 0.538 &  4.33& 0.512 & 4.37 & 0.591 & 4.16 & 0.683\tabularnewline
 & 32 & 3.6 & 0.514 & 3.61  & 0.496 & 3.68 & 0.579 & 3.46 & 0.679\tabularnewline
 & 16 & - & - & 2.82 & 0.217 & 2.9 & 0.308 & 2.79 & 0.513\tabularnewline

\toprule[1.2pt]

\multirow{1}{0.15\textwidth}{\centering{} S. Kim et al.\cite{kim2021zero} } 
&  & & & & & & & FP32 & 54.64 \tabularnewline
& - & - & - &- & - &-&-& 6.58 & 54.52\tabularnewline
\midrule

\multirow{1}{0.15\textwidth}{\centering{} Q-YOLO\cite{wang2023q} } 
&  & & & & & FP32 &  57.1 &-&- \tabularnewline
& - & - & - &- & - & 8 &  56.9&-&-\tabularnewline
& - & - & - &- & - & 4 &  26.2&-&-\tabularnewline

\bottomrule[1.5pt]
\end{tabular*}

\label{table:comparison-with-competitors-per-channel_yolo}
\end{table*}


\paragraph{Object detection}
We further evaluate the proposed VLCQ and its variants on the object detection challenge. The evaluations are carried out on the public Pytorch implementation of the SSD-Lite version with a MobileNetV2 backbone, on the Pascal VOC object detection challenge, performance is measured using mean average precision (mAP).
{ In addition to SSD-Lite, several evaluations have been carried out on different versions of YOLO model \cite{yolo11_ultralytics,farhadi2018yolov3,varghese2024yolov8}.} 
Table \ref{table:comparison-with-competitors-per-channel_pascal_voc_obj_detect} compares the results of the mAP score of quantized models between PWLQ and VLCQ along with its variants for per-channel quantization. VLCQ, VLCQ Min-Max, VLCQ Min-Max SS, VLCQ SS achieve better results for 256, 128, 64, 32 levels quantization compared to PWLQ. In addition, comparing Min-Max adjust, scale selection and their combination Min-Max SS show promising results in improving the performance for low bit-rate. For instance, VLCQ Min-Max SS improved performance for 8 levels quantization with magnitude of 20\% when compared to VLCQ Min-Max. Also VLCQ Min-Max SS improved performance for 16 levels quantization by magnitude of 8\% when compared with VLCQ SS, 10\% when compared with VLCQ. \\
{
Table \ref{table:comparison-with-competitors-per-channel_yolo} compares the results of the mAP score of quantized models between Q-YOLO, and S. Kim et al. \cite{kim2021zero} and VLCQ along with its variants for per-channel quantization. VLCQ, VLCQ Min-Max, VLCQ Min-Max SS, VLCQ SS achieve better results for 256, 128, 64, 32, 16 levels quantization compared to Q-YOLO, and \cite{kim2021zero}. VLCQ, VLCQ Min-Max, and VLCQ Min-Max SS achieves the same FP32 performance for a bit-rate ranging 5-6. For low bit-rate VLCQ Min-Mac SS outperforms other methods. }
\\ \\ 
Although the current proposed methodology, VLCQ and its variants, show impressive results in various tasks, it suffers from some limitations. For instance, in scale selection and min-max adjustment, the search space is fixed. The search space should be dynamic according to bit-rate, and model architecture. During our experiments for higher bit-rate it requires a small search space for low bit-rate. So, to achieve good results, the search space needs to be increased. The metric to choose the best scale or minimum and maximum values needs to be adjusted to investigate which p-value in lp-norm to choose, currently we are fixing it to 2 (MSE). Another limitation is the choice of break-point currently we are limiting to zero, even for the distribution which lies entirely to the left or right. This might need more investigation to find the best break point to start quantization from. Another limitation is performance degradation in case of outlier presence. Although Scale selection and Min-Max could mitigate this case in per-channel quantization as in the case of semantic segmentation in table \ref{table:comparison-with-competitors-per-channel_pascal_voc_ss}, it wasn't able to achieve relatively good performance in case of per-tensor quantization. However, if search space in per-tensor quantization was to be large and the presence of calibration data was to be used to choose the best min-max or scale. Performance would drastically improve.

\section{Conclusion}
\label{sec:Conclusion}
In this paper, we introduced VLCQ: post-training quantization with variable-length encoding; A data-free method for quantization, which requires no access to the original full dataset. VLCQ leverages from the normal distribution of weight, by dividing weights into two regions, where each region is assigned an equal number of bins. In addition to that, we leverage this distribution to further compress weights achieving low bit-width, thus low memory footprint. VLCQ delivers near-original model accuracy for nearly every model when evaluated, and even competes with more complex training-based methods. Furthermore, The suggested framework is efficient and free of hyperparameters, and it may be readily implemented and integrated into existing network quantization libraries.

 \section*{Acknowledgment}
\label{sec:Acknowledgement}
This paper is based upon work supported by Science and Technology Development Fund (STDF) under grant 45164.

\bibliographystyle{unsrt}  
\bibliography{references}

\end{document}